\documentclass[letterpaper]{article}
\usepackage[preprint]{aaai2027}
\usepackage[hyphens]{url}
\usepackage{graphicx}
\usepackage{natbib}
\usepackage{caption}
\usepackage{booktabs}
\usepackage{multirow}
\usepackage{array}
\usepackage{makecell}
\usepackage{pifont}
\usepackage[most]{tcolorbox}

\newcommand{\cmark}{\ding{51}}
\newcommand{\xmark}{\ding{55}}

\title{FitAQA: A Benchmark of Fitness Action Quality Assessment \\for Multimodal Large Language Models}
\author{
    Kaili Zheng\textsuperscript{\rm 1}\equalcontrib,
    Kaiwen Wang\textsuperscript{\rm 1}\equalcontrib,
    Xun Zhu\textsuperscript{\rm 1},
    Qingyuan Yang\textsuperscript{\rm 2},
    Chenyi Guo\textsuperscript{\rm 1}\corresponding,
    Ji Wu\textsuperscript{\rm 1, \rm 3, \rm 4}\corresponding
}
\affiliations{
    \textsuperscript{\rm 1}Department of Electronic Engineering, Tsinghua University\\
    \textsuperscript{\rm 2}Xinjiang Region Sports Science Research Center\\
    \textsuperscript{\rm 3}College of AI, Tsinghua University\\
    \textsuperscript{\rm 4}Beijing National Research Center for Information Science and Technology\\
    \{guochy, wuji\_ee\}@mail.tsinghua.edu.cn
}

\begin{document}

\maketitle

\begin{abstract}
Fitness Action Quality Assessment (AQA) is important for intelligent sports training, yet the capabilities of Multimodal Large Language Models (MLLMs) in this setting remain underexplored. Existing benchmarks rely on action-specific annotation schemes and focus primarily on final assessment outputs, offering limited insight into how models assess exercise quality.
We introduce \textbf{FitAQA}, a systematic benchmark for evaluating MLLMs in fitness AQA, containing 2,219 videos and 5,512 QA instances across 30 bodyweight exercises.
In collaboration with experts in sports science, we develop a unified form error taxonomy that defines 38 recurring form errors within six complementary quality dimensions: alignment, symmetry, stability, coordination, tempo, and completeness. This taxonomy provides a shared assessment framework across different exercises.
FitAQA further formulates three evaluation tasks: perception for recognizing relevant visual evidence, judgement for combining that evidence with domain knowledge to assess execution correctness, and temporal grounding for localizing form errors over time. 
Extensive evaluation shows that current MLLMs still struggle to assess exercise quality comprehensively and localize form errors precisely. 
Controlled experiments further indicate that visual perception is a key bottleneck, as judgement performance improves substantially when ground-truth perceptual evidence is provided. The dataset is available at \url{https://huggingface.co/datasets/Kelly0510/FitAQA}.
\end{abstract}

\section{Introduction}

Action Quality Assessment (AQA) aims to evaluate the quality of human actions. Unlike conventional action recognition, which focuses on identifying \emph{what} action is being performed~\cite{carreira2017quo,salehi2024actionatlas}, AQA emphasizes \emph{how well} the action is executed~\cite{parmar2017learning,parmar2019and}. This capability is particularly important in fitness scenarios, where subtle differences in posture, motion trajectory, and execution rhythm can directly affect training effectiveness and movement safety~\cite{wilk2021influence,bonilla2022exercise}. With the growing popularity of online and home-based fitness training, automatic fitness AQA has become increasingly important for corrective feedback and personalized guidance. Consequently, fitness AQA has attracted increasing attention in recent years.

Meanwhile, Multimodal Large Language Models (MLLMs)~\cite{liu2023visual,hurst2024gpt,maaz2024video,LLaVA-OneVision-2} have shown growing capabilities in general visual understanding. Unlike traditional methods that predict scores~\cite{tang2020uncertainty,yu2021group} or labels for predefined action types~\cite{hulsmann2018classification,jaiswal2025real}, MLLMs enable action-agnostic assessment with free-form responses, making them a promising foundation for flexible fitness coaching systems. Despite this potential, their effectiveness for fitness AQA remains underexplored.

\begin{figure*}[htbp]
  \centering
  \includegraphics[width=\linewidth]{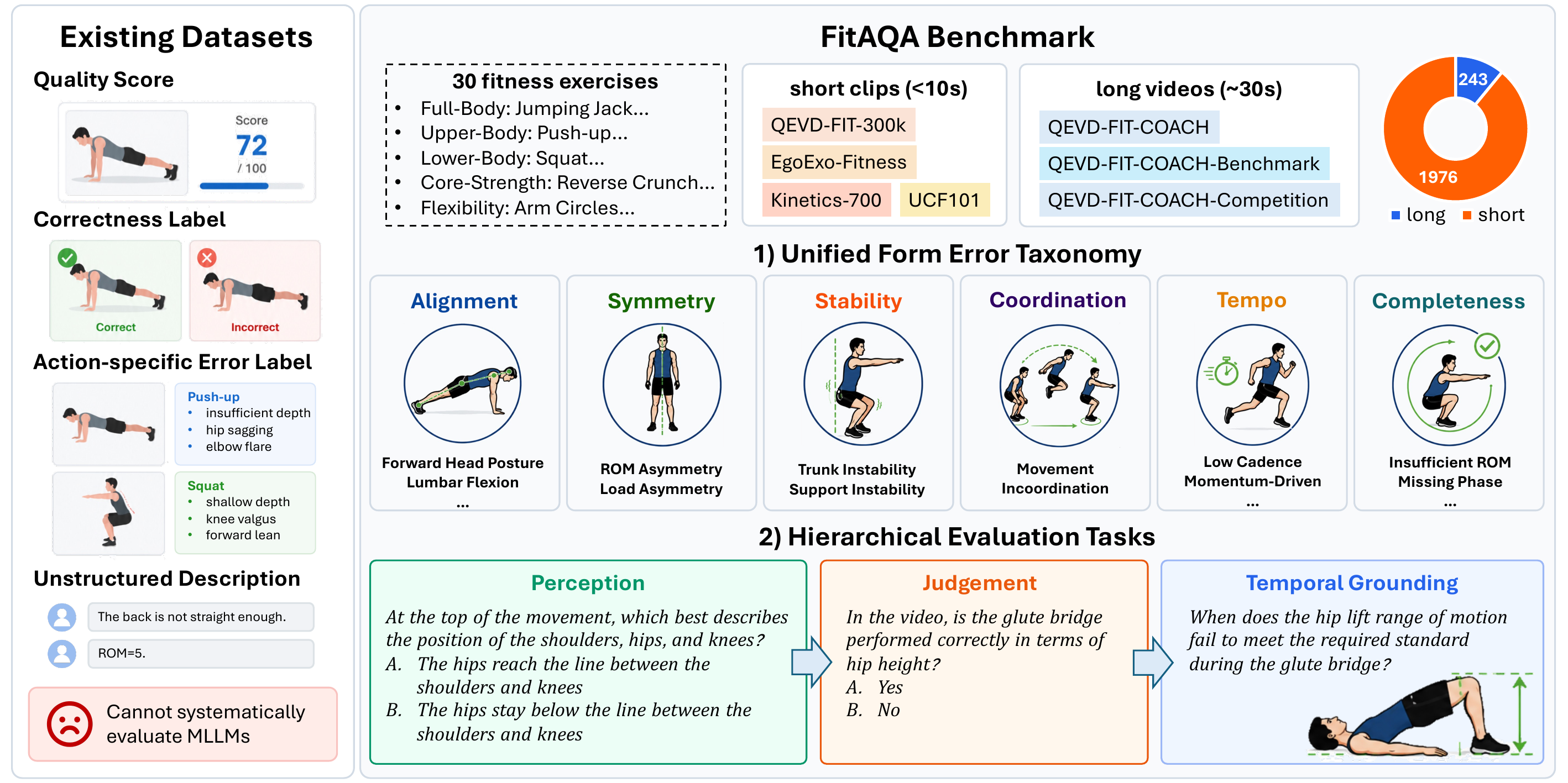}
  \caption{Overview of FitAQA. (a) Unified form error taxonomy enables cross-exercise evaluation through an action-agnostic label space. (b) Hierarchical evaluation tasks decompose fitness AQA into perception, judgement, and temporal grounding.}
  \label{fig:teaser}
\end{figure*}

Unlike general video understanding tasks, fitness AQA requires models to recognize the details of body posture and motion dynamics that determine execution quality, and to reason with exercise-specific knowledge of proper form standards, common form errors, and basic biomechanical principles.
Existing video understanding benchmarks for MLLMs~\cite{li2024mvbench,fu2025video,cai2025humanvideo} primarily evaluate semantic understanding of video content and offer limited evidence of models' sensitivity to such quality-relevant posture and motion cues.
Meanwhile, traditional fitness AQA datasets are insufficient for systematically evaluating MLLMs. They rely on scalar quality scores~\cite{parmar2019action}, action-specific error labels~\cite{youssef2022analysis}, or unstructured descriptions~\cite{li2024egoexo}. 
Consequently, results are difficult to compare across datasets and exercises. Moreover, evaluation of final outputs alone offers limited insight into why models fail.

To bridge this gap, we introduce \textbf{FitAQA}, a benchmark for systematically evaluating MLLMs in fitness AQA, as illustrated in Figure~\ref{fig:teaser}. FitAQA contains 2,219 videos and 5,512 QA instances covering 30 bodyweight exercises, including both short clips and longer videos. In collaboration with experts in sports science, we develop a \textbf{unified form error taxonomy}. Rather than defining a separate label space for each exercise, the taxonomy defines 38 recurring posture and motion errors spanning six quality dimensions: alignment, symmetry, stability, coordination, tempo, and completeness. During annotation, annotators select all applicable form-error labels and describe how each selected error manifests in the video. The resulting annotations combine an action-agnostic label space with video-specific descriptions of observable evidence. FitAQA further defines three \textbf{hierarchical evaluation tasks}: perception, judgement, and temporal grounding. Paired perception and judgement questions target the same quality aspect. Perception evaluates whether models recognize the relevant visual evidence, whereas judgement further requires them to combine the observed evidence with domain knowledge to assess execution correctness. Temporal grounding extends the evaluation to longer videos by requiring models to locate when a described form error occurs. Together, these tasks provide a hierarchical view of the capabilities involved in fitness AQA, spanning visual perception, quality judgement, and temporal localization.

We conduct an extensive evaluation of current MLLMs on FitAQA. The results show that most models perform on par with or only marginally better than simple baselines, and none achieves consistently strong performance across all six quality dimensions. Perception accuracy remains low across models, while providing ground-truth perceptual evidence substantially improves judgement, identifying visual perception as a key bottleneck. Temporal grounding also remains challenging, with performance degrading markedly under strict overlap thresholds, indicating limited ability to precisely localize form errors. Taken together, these findings suggest that capabilities demonstrated on general video understanding benchmarks do not readily transfer to recognizing and localizing the quality-relevant posture and motion evidence required for fitness AQA.

Our contributions are summarized as follows:
\begin{itemize}
  \item We introduce FitAQA, a systematic benchmark for evaluating MLLMs in fitness AQA, comprising 2,219 videos and 5,512 QA instances across 30 bodyweight exercises.
  \item We develop a unified form error taxonomy that defines 38 recurring errors across six quality dimensions, providing a shared label space for cross-exercise evaluation. We further formulate hierarchical evaluation tasks that decompose fitness AQA into perception, judgement, and temporal grounding.
  \item We conduct an extensive evaluation of current MLLMs, revealing limited recognition and temporal localization of quality-relevant posture and motion evidence. Controlled experiments further identify visual perception as a key bottleneck for judgement.
\end{itemize}

\section{Related Work}
\subsubsection{Traditional Action Quality Assessment}

Traditional AQA methods often formulate action assessment as score regression or classification problems. Score regression is primarily adopted in competitive sports such as diving~\cite{parmar2017learning,parmar2019and}, figure skating~\cite{xu2019learning}, gymnastics~\cite{shao2020finegym,dong2024lucidaction}, artistic swimming~\cite{zhang2023logo}, and skiing~\cite{zhang2025fineskiing}, where professional judging systems make quality scores naturally available as supervision. In contrast, classification-based AQA is more common in fitness and rehabilitation scenarios, where scalar score supervision is not readily available. Some works~\cite{liao2020deep} perform correctness classification to determine whether an exercise is executed properly, while others~\cite{parmar2022domain,marusic2025skeleton} define action-specific error labels and perform multi-label classification. However, both regression and classification provide limited feedback for users. To improve interpretability, several traditional methods further generate explanatory feedback. For example, AIFit produces rule-based natural language feedback~\cite{fieraru2021aifit}. Besides, Zhao et al.~\cite{zhao20223d} and Zheng et al.~\cite{zheng2023skeleton} provide skeleton-based feedback through weak supervision. More recently, LLM-FMS~\cite{xing2025llm} and Zhang et al.~\cite{zhang2026exercise} use LLMs to interpret structured pose-derived cues for explainable exercise assessment. Overall, traditional AQA methods still largely rely on task-specific models, predefined output spaces, or rule-based cues, limiting their application in open-ended assessment.

\subsubsection{MLLM-based Action Quality Assessment}

Recent studies have begun to adapt AQA to MLLMs by redesigning task formulations and supervision signals for multimodal reasoning and feedback generation. FLEX~\cite{yin2025flex} constructs multimodal and multiview fitness data with structured annotations of action steps, error types, and corrective feedback, enabling quality-related video question answering. Qi et al.~\cite{qi2025explainable} introduce chain-of-thought annotations for explainable action form assessment. Panchal et al.~\cite{panchal2024say} study live fitness coaching, where models need to decide both what feedback to provide and when to provide it, while BioCoach~\cite{ji20263d} further grounds coaching feedback in 3D pose and biomechanical cues. Dibenedetto et al.~\cite{dibenedetto2025fine} fine-tune a large multimodal model for fitness error detection and temporal localization.
Despite these advances, existing MLLM-based AQA evaluations remain limited in their ability to diagnose model capabilities. Some works still evaluate MLLMs with traditional AQA metrics. Dibenedetto et al.~\cite{dibenedetto2025fine} and Freitas et al.~\cite{freitas2026can} assess models through score regression, correctness classification, or exercise-specific error recognition under existing dataset definitions. Other works focus on generated feedback or explanations. FLEX~\cite{yin2025flex}, Qi et al.~\cite{qi2025explainable}, Panchal et al.~\cite{panchal2024say}, and BioCoach~\cite{ji20263d} mainly evaluate generated answers, explanations, or coaching feedback using generic text-generation metrics or LLM-as-Judge scores. These evaluations measure whether the final output matches annotations, but provide limited insight into why models fail.

\section{FitAQA Benchmark}
\begin{figure*}[t!]
  \centering
  \includegraphics[width=\textwidth]{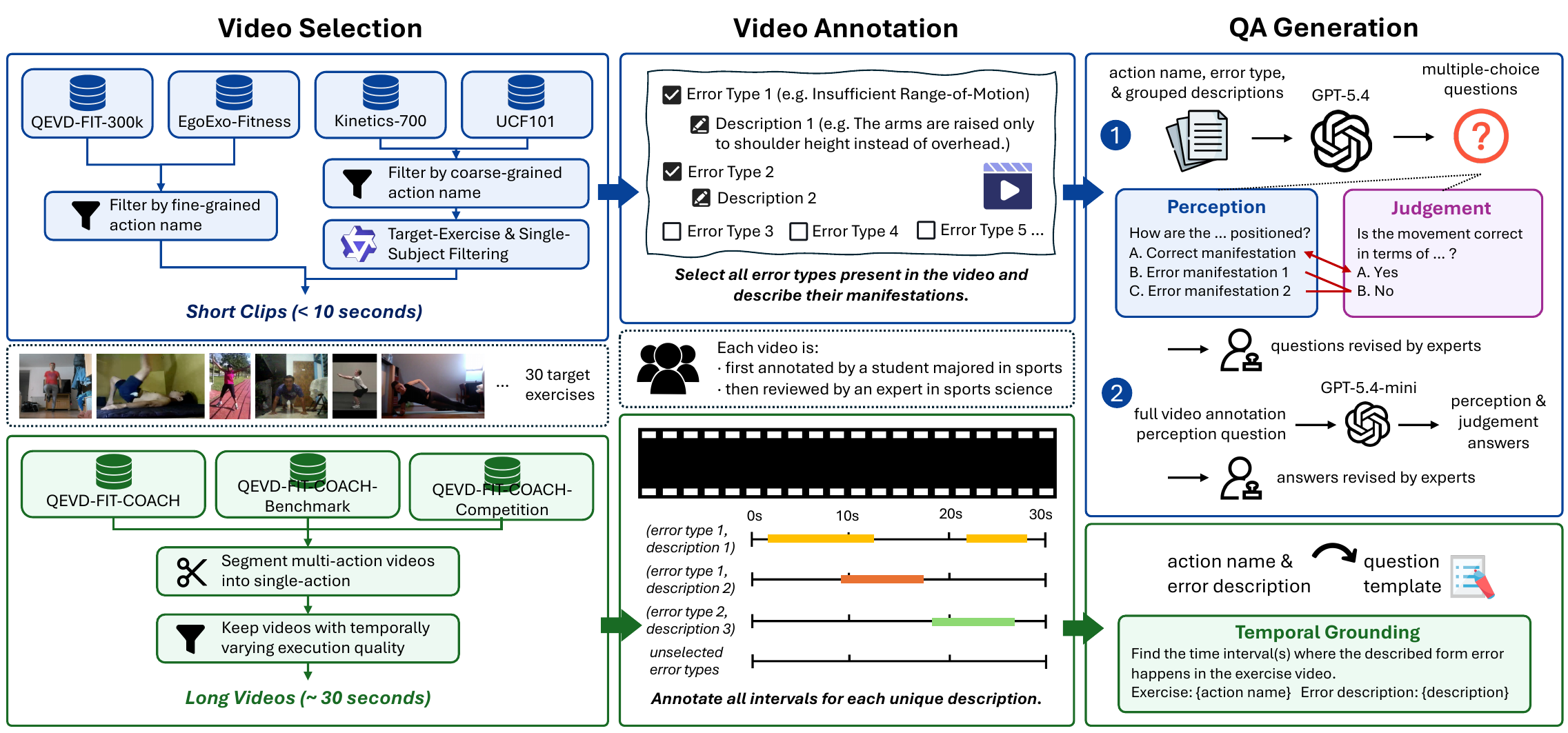}
  \caption{Construction pipeline of FitAQA.}
  \label{fig:construction_pipeline}
\end{figure*}

This section describes the construction pipeline of FitAQA, summarized in Figure~\ref{fig:construction_pipeline}, and presents the resulting dataset statistics. Further details are provided in the Appendix.

\subsection{Unified Form Error Taxonomy}
Although bodyweight exercises differ in target muscles and movement patterns, experts evaluate their execution along a shared set of quality dimensions. This motivates the unified form error taxonomy in FitAQA, which captures recurring form errors across exercises.
The taxonomy was developed in collaboration with experts in sports science. 
Before annotation, two experts formulated an initial taxonomy based on theoretical knowledge and practical experience in bodyweight fitness training, focusing on posture and motion errors that commonly recur across bodyweight exercises. 
During annotation, the taxonomy was further refined in an iterative manner: annotators reported ambiguous or uncovered cases, and the experts updated the taxonomy when a recurring error could not be cleanly assigned to an existing type.
The final taxonomy defines 38 form errors across six quality dimensions.
\textit{Alignment} covers deviations in joint and body-segment configuration, \textit{Symmetry} captures left-right discrepancies, and \textit{Stability} concerns failures to maintain controlled support and trunk position.
\textit{Coordination} captures temporal relationships among body parts, \textit{Tempo} concerns movement rhythm and cadence, and \textit{Completeness} assesses whether the required movement phases and range of motion are fully executed.
The full taxonomy is provided in the Appendix.

\subsection{Video Selection}
FitAQA covers 30 common bodyweight exercises across full-body cardio, upper-body strength, lower-body strength, core strength, and flexibility.
To support the three evaluation tasks, we construct two video subsets: short clips for perception and judgement, and longer videos for temporal grounding.
The short-clip subset is drawn from QEVD-FIT-300k~\cite{panchal2024say}, EgoExo-Fitness~\cite{li2024egoexo}, Kinetics-700~\cite{carreira2019short}, and UCF101~\cite{soomro2012ucf101}, combining fitness-specific sources with generic action-recognition videos to improve diversity in recording conditions.
We map source action labels to the 30 target exercises and filter generic videos to retain single-subject clips that match the target exercises.
The long-video subset comes from the longer-video subsets of QEVD~\cite{panchal2024say}, including QEVD-FIT-COACH, QEVD-FIT-COACH-Benchmark, and QEVD-FIT-COACH-Competition.
We segment workout videos into single-action clips using source-provided timestamps and retain only those with temporal variation in execution quality. This design requires models to localize when a form error occurs rather than merely recognize the action.

\subsection{Video Annotation}
Video annotation in FitAQA is conducted based on the unified form error taxonomy. 
Three annotators, all students majoring in sports, participated in the annotation process. 
Each video was first annotated by one annotator and then reviewed by one expert. 
For the short-clip subset, annotators selected all visible form errors in each video. 
For every selected form error, annotators further wrote a detailed description for this specific video. 
These descriptions preserve video-specific visual evidence while keeping the label space unified. 
For the long-video subset, annotators additionally marked the temporal intervals where each error occurred. 
Since the same form-error type may have multiple descriptions within a video when it manifests differently over time, annotators marked all corresponding intervals for each unique description.

\subsection{Question and Answer Generation}
Using the resulting video annotations, we construct QA instances for the three hierarchical evaluation tasks in FitAQA: perception, judgement, and temporal grounding.

For the perception and judgement tasks, we group short clips by action type and form-error type and collect the video-specific descriptions within each group.
We provide the deduplicated descriptions to GPT-5.4~\cite{singh2025openai} and prompt it to generate an initial perception and judgement question pair for each action-error group.
The perception question asks which observable manifestation of the target aspect is present. All options are phrased as neutral descriptions of visual manifestations and do not explicitly label the execution as correct or erroneous. Option A corresponds to the visual manifestation associated with correct execution, while the remaining options correspond to different error manifestations.
The paired judgement question asks whether the same aspect is executed correctly, with binary Yes / No options.
By holding the assessed aspect fixed, the pair separates recognition of visual evidence from judgement that combines this evidence with domain knowledge.
Conditioning question generation on the annotated descriptions grounds the options primarily in visual variations observed in the dataset and avoids wording tied to any single video.
Experts then review and revise the question pairs to ensure that each perception option maps unambiguously to its corresponding judgement answer and to merge pairs assessing the same quality aspect.

For each revised perception question, we use GPT-5.4-mini to infer an answer for every video of the corresponding action type based on its form-error labels and descriptions.
The model is allowed to abstain when the annotation does not provide sufficient evidence to select a unique option.
We then automatically derive the judgement answer from the perception answer: option A maps to \textit{Yes}, non-A options map to \textit{No}, and abstained perception answers remain abstained.
Experts then inspect the corresponding videos to verify the inferred answers, correcting the selected option or marking an instance as unanswerable when necessary.
After expert verification, we downsample the QA instances using mixed-integer linear programming (MILP) to reduce answer-option imbalance while preserving coverage of videos, unique questions, and option buckets, as detailed in the Appendix.

For the temporal grounding task, we construct questions directly from the long-video annotations using the template:
\begin{quote}
\small
Find the time interval(s) where the described form error happens in the exercise video.\\
Exercise: \{action name\}\quad Error description: \{description\}
\end{quote}
We construct one question for each unique combination of video, form error, and description, requiring models to localize all intervals in which the described error occurs.

\subsection{Dataset Statistics}

\begin{figure*}[t!]
  \centering
  \includegraphics[width=\textwidth]{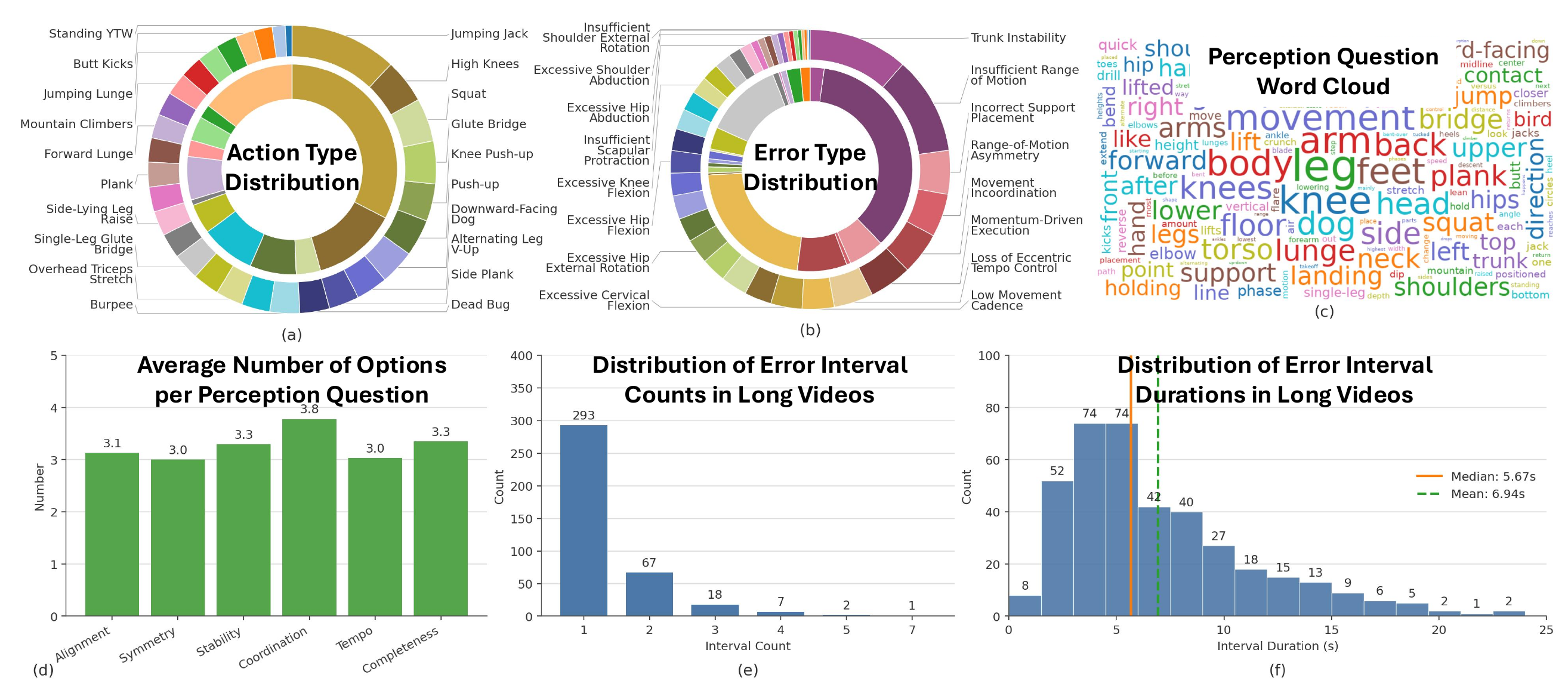}
  \caption{Dataset statistics. In (a) and (b), outer and inner rings denote the short-clip and long-video subsets.}
  \label{fig:dataset_stats}
\end{figure*}

Figure~\ref{fig:dataset_stats} summarizes the dataset statistics. FitAQA contains 2,219 videos, including 1,976 short clips for perception and judgement and 243 long videos for temporal grounding.
The benchmark contains 288 unique perception and judgement question pairs. Instantiating these questions with compatible short clips yields 2,562 perception and 2,562 judgement QA instances. The temporal grounding subset contains 388 QA instances, resulting in 5,512 QA instances overall.
The form error distribution is naturally long-tailed, reflecting the fact that some errors, such as trunk instability and insufficient range of motion, occur across many exercises, while others are specific to particular movement patterns. The average number of options per perception question ranges from 3.0 to 3.8 across the six quality dimensions. Most temporal grounding instances contain a single target interval, while 95 require localizing multiple intervals. The interval-duration distribution exhibits substantial diversity, covering both transient and sustained form errors.

\subsection{Dataset Quality Validation}
To examine annotation consistency, we draw a stratified random sample of the final QA instances. Each expert re-annotates sampled perception and temporal grounding instances previously reviewed by the other expert. Together, the two subsets cover 10.0\% of the full dataset, including 257 perception instances and their 257 matched judgement instances, along with 39 temporal grounding instances. Exact-answer agreement for perception reaches 98.4\%, with the mapped judgement answers inheriting the same agreement. For temporal grounding, the independently annotated temporal intervals achieve an mIoU of 97.3\%. These cross-check results provide evidence of high annotation consistency.

\section{Experiments}

\subsection{Evaluated Models}

We evaluate a broad range of MLLMs, covering both open-source model families and closed-source systems.
For open-source models, we include VideoLLaMA3~\cite{zhang2025videollama}, Qwen3-VL~\cite{bai2025qwen3}, Qwen3.5~\cite{qwen35}, InternVL3.5~\cite{wang2025internvl3}, and Gemma-4~\cite{gemma4}. For the Qwen3.5 series, we evaluate both non-thinking and thinking modes, while for the other open-source families, we use their instruct models. For closed-source systems, we evaluate Gemini and GPT-series models, including Gemini-3.1-pro-preview~\cite{gemini31pro}, GPT-5.4, and GPT-5.5.
For temporal grounding, we further evaluate specialized temporal grounding models, including Grounded-VideoLLM~\cite{wang2024grounded}, VideoMind~\cite{liu2025videomind} and TimeLens~\cite{zhang2025timelens}.

Open-source models are evaluated on two NVIDIA A800 GPUs. We use each model's official video preprocessing pipeline and default decoding settings, running inference with vLLM~\cite{kwon2023efficient} when compatible and with the corresponding official Transformers implementation~\cite{wolf2020transformers} otherwise. Closed-source models are evaluated through their official APIs, with the requested frame rate set to 2 fps. UCF101 samples are instead evaluated at their native frame rate of 1 fps for all models.

\subsection{Evaluation Protocol}

For perception, we report accuracy and question-macro accuracy (Q-MAcc), which first computes accuracy for each unique perception question and then averages across questions, giving every question equal weight. 
For judgement, we report recall, precision, and F1, treating the answer indicating a form error as the positive class.
For perception and judgement, we include uniform-random and always-correct baselines. The former samples uniformly from the available options. The latter always selects the correct-execution answer and is included because prior work has observed that MLLMs tend to overpredict correct execution~\cite{freitas2026can}.
For temporal grounding, we report recall at IoU thresholds of 0.3, 0.5, and 0.7, together with mean IoU (mIoU), adapting standard temporal grounding metrics~\cite{wang2024grounded,zhang2025timelens} to the multi-interval setting as detailed in the Appendix. The whole-video baseline predicts the entire video as the target interval.
All main results are averaged across three runs.

\subsection{Main Results}

\begin{table*}[t!]
  \centering
  \small
  \begin{tabular}{lc|cc|ccc|cccc}
    \toprule
    \multirow{2}{*}{\textbf{Model}} &
    \multirow{2}{*}{\textbf{Think}} &
    \multicolumn{2}{c|}{\textbf{Perception}} &
    \multicolumn{3}{c|}{\textbf{Judgement}} &
    \multicolumn{4}{c}{\textbf{Temporal Grounding}} \\
    & & \textbf{Accuracy} & \textbf{Q-MAcc} & \textbf{Recall} & \textbf{Precision} & \textbf{F1} & \textbf{R@0.3} & \textbf{R@0.5} & \textbf{R@0.7} & \textbf{mIoU} \\
    \midrule
    \multicolumn{11}{l}{\textit{Simple Baselines}} \\
    Uniform Random & -- & 33.6 & 35.7 & 50.0 & 58.9 & 54.1 & -- & -- & -- & -- \\
    Always Correct & -- & 41.1 & 43.2 & 0.0 & 0.0 & 0.0 & -- & -- & -- & -- \\
    Whole Video & -- & -- & -- & -- & -- & -- & 25.5 & 4.9 & 0.0 & 23.2 \\
    \midrule
    \multicolumn{11}{l}{\textit{Temporal Grounding Models}} \\
    Grounded-VideoLLM-Phi3.5 & -- & -- & -- & -- & -- & -- & 11.9 & 4.4 & 0.5 & 13.1 \\
    VideoMind & -- & -- & -- & -- & -- & -- & 33.0 & 14.9 & 4.9 & 22.8 \\
    TimeLens & -- & -- & -- & -- & -- & -- & 28.6 & 16.2 & 7.5 & 22.0 \\
    \midrule
    \multicolumn{11}{l}{\textit{Open-source Models}} \\
    VideoLLaMA3-7B & \xmark & 38.4 & 40.8 & 18.9 & 59.6 & 28.7 & 27.3 & 12.1 & 4.9 & 18.4 \\
    InternVL3.5-8B & \xmark & 40.1 & 43.7 & 19.5 & 64.2 & 29.9 & 19.8 & 3.6 & 0.5 & 17.3 \\
    InternVL3.5-30B-A3B & \xmark & 41.7 & 44.6 & 10.3 & 65.1 & 17.7 & 15.2 & 5.2 & 1.3 & 12.1 \\
    Gemma-4-31B & \xmark & \textbf{43.4} & \textbf{46.3} & 17.6 & 65.8 & 27.8 & 35.6 & 16.5 & 5.4 & 24.9 \\
    Qwen3-VL-8B & \xmark & 41.3 & 44.2 & 16.9 & 64.9 & 26.8 & 12.4 & 3.4 & 0.3 & 10.1 \\
    Qwen3-VL-30B-A3B & \xmark & 41.4 & 44.3 & 12.1 & 64.8 & 20.3 & 32.7 & 15.7 & 7.5 & 24.3 \\
    Qwen3.5-9B & \xmark & 41.3 & 44.6 & 35.8 & 63.0 & 45.7 & 36.1 & 19.1 & 7.0 & 25.7 \\
    Qwen3.5-9B & \cmark & 41.1 & 43.9 & 28.9 & 64.5 & 39.9 & 42.3 & 22.7 & 10.3 & 30.4 \\
    Qwen3.5-27B & \xmark & 42.5 & 45.2 & \textbf{37.4} & 63.4 & \textbf{47.1} & 51.3 & \textbf{35.1} & \textbf{19.3} & \textbf{37.6} \\
    Qwen3.5-27B & \cmark & 42.6 & 45.7 & 31.3 & \textbf{66.7} & 42.6 & \textbf{52.3} & 30.2 & 17.0 & 36.6 \\
    Qwen3.5-35B-A3B & \xmark & 42.4 & 45.6 & 22.5 & 64.5 & 33.4 & 43.3 & 26.8 & 13.7 & 30.5 \\
    Qwen3.5-35B-A3B & \cmark & 42.5 & 45.9 & 27.7 & 65.7 & 39.0 & 45.9 & 25.8 & 12.9 & 31.7 \\
    \midrule
    \multicolumn{11}{l}{\textit{Closed-source Models}} \\
    Gemini-3.1-pro-preview & \cmark & 50.7 & \textbf{54.3} & \textbf{80.6} & 65.2 & \textbf{72.1} & 62.4 & \textbf{43.8} & 25.8 & 44.0 \\
    GPT-5.4 & \cmark & 47.6 & 51.3 & 38.8 & \textbf{72.3} & 50.5 & 51.3 & 28.5 & 18.6 & 36.9 \\
    GPT-5.5 & \cmark & \textbf{51.2} & 54.1 & 48.3 & 71.9 & 57.8 & \textbf{67.3} & 42.4 & \textbf{27.0} & \textbf{47.5} \\
    \bottomrule
  \end{tabular}
  \caption{Overall results on the three tasks of FitAQA. R@$x$ denotes Recall@$x$. All values are reported as percentages.}
  \label{tab:main_results}
\end{table*}

\begin{figure*}[t!]
  \centering
  \includegraphics[width=\textwidth]{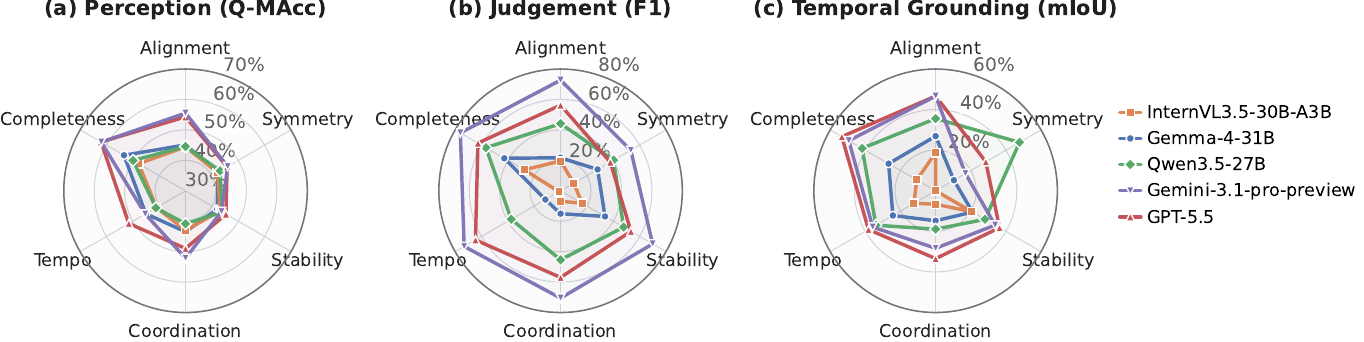}
  \caption{Model performance across six quality dimensions on the three tasks of FitAQA.}
  \label{fig:dimension_radar}
\end{figure*}

Table~\ref{tab:main_results} reports results across three tasks. Overall, the performance of current MLLMs on FitAQA remains limited. Closed-source models generally outperform the evaluated open-source models, which achieve only modest gains over simple baselines and fall below them on some metrics.
Perception is particularly challenging. Gemini-3.1-pro-preview achieves the highest Q-MAcc of 54.3\%, while the best open-source result is 46.3\% from Gemma-4-31B, only 3.1 points above the always-correct baseline. These results indicate difficulty in identifying the quality-relevant posture and motion evidence in the videos. For judgement, Gemini-3.1-pro-preview achieves the highest F1 of 72.1\%. Among all other MLLMs, recall is at most 48.3\% and lower than precision. This pattern aligns with prior observations that MLLMs tend to overpredict correct execution and miss form errors~\cite{freitas2026can}. All evaluated open-source models fall below the uniform-random baseline in F1, and thinking mode provides no consistent improvement across tasks.
Temporal grounding performance declines sharply as the IoU threshold becomes stricter. GPT-5.5 performs best with 27.0\% Recall@0.7 and 47.5\% mIoU, while Qwen3.5-27B is the strongest open-source model with 19.3\% Recall@0.7 and 37.6\% mIoU. None of the specialized temporal grounding models surpasses the whole-video baseline in mIoU. 
These results suggest that current MLLMs struggle to recognize quality-relevant evidence, detect form errors, and localize them precisely.

\begin{figure*}[t!]
  \centering
  \begin{minipage}[t]{0.62\textwidth}
    \vspace{0pt}
    \centering
    \small
    \setlength{\tabcolsep}{5pt}
    \begin{tabular}{lccc}
      \toprule
      \textbf{Setting} & \textbf{Gemma-4-31B} & \textbf{Qwen3.5-27B} & \textbf{GPT-5.5} \\
      \midrule
      Baseline         & 17.6 / 65.8 / 27.8 & 37.4 / 63.4 / 47.1 & 48.3 / 71.9 / 57.8 \\
      Question-only    & 17.9 / 67.8 / 28.4 & 21.7 / 64.9 / 32.5 & 48.0 / 72.1 / 57.6 \\
      Model-perception & 11.8 / 72.4 / 20.3 & 16.2 / 71.8 / 26.4 & 45.7 / 73.0 / 56.2 \\
      GT-perception    & 63.2 / 96.3 / 76.3 & 65.6 / 97.0 / 78.2 & 90.3 / 99.0 / 94.4 \\
      GT-perception, no video & 68.9 / 99.1 / 81.3 & 66.8 / 98.9 / 79.8 & 92.6 / 99.2 / 95.8 \\
      \bottomrule
    \end{tabular}
    \captionof{table}{Judgement performance under different perception oracle settings. Each cell reports error \textit{recall / precision / F1} as percentages.}
    \label{tab:oracle_perception}
  \end{minipage}
  \hfill
  \begin{minipage}[t]{0.35\textwidth}
    \vspace{0pt}
    \centering
    \includegraphics[width=\linewidth]{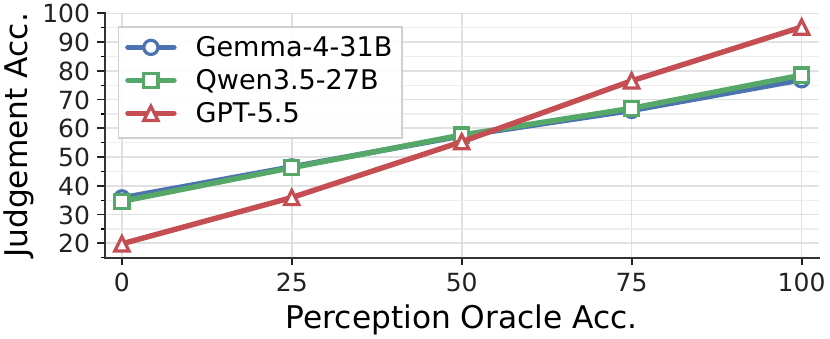}
    \captionof{figure}{Judgement accuracy as a function of perception oracle accuracy.}
    \label{fig:oracle_perception}
  \end{minipage}
\end{figure*}

\begin{figure*}[t!]
  \centering
  \includegraphics[width=\textwidth]{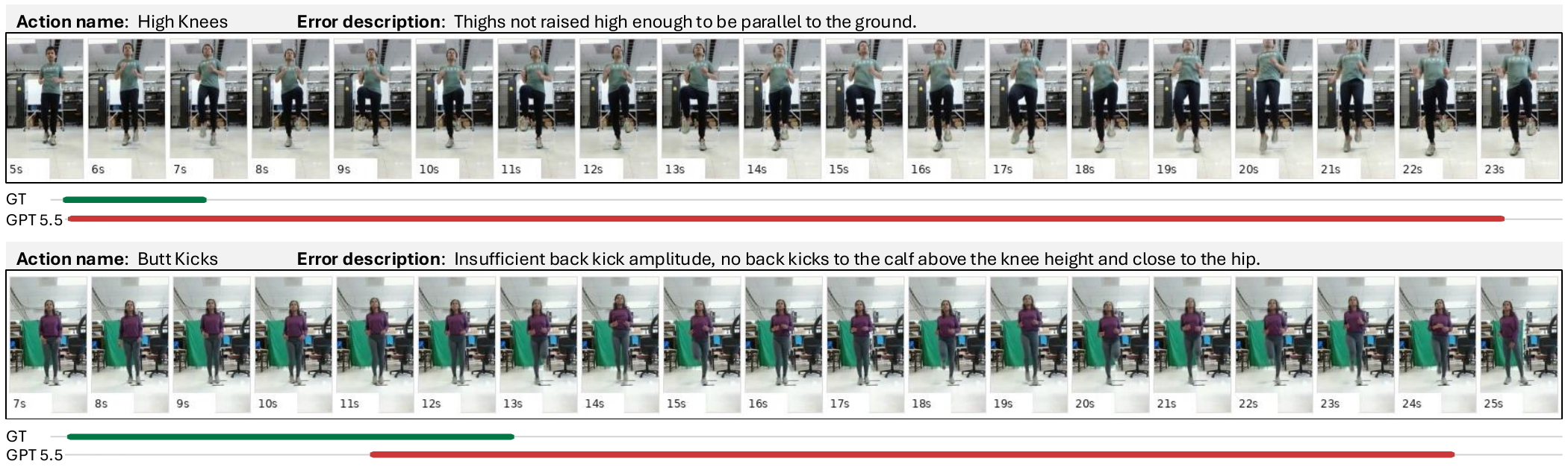}
  \caption{Temporal grounding failure cases. Green bars denote ground truths and red denote predictions of GPT-5.5.}
  \label{fig:grounding_vis}
\end{figure*}

Figure~\ref{fig:dimension_radar} shows how each model performs across the six quality dimensions. In perception, GPT-5.5 and Gemini-3.1-pro-preview achieve similar Q-MAcc but exhibit different dimension profiles, most notably in the stronger tempo performance of GPT-5.5. The judgement advantage of Gemini-3.1-pro-preview extends across all six dimensions. In temporal grounding, Qwen3.5-27B achieves the highest mIoU on symmetry despite trailing the two closed-source models overall. Although individual models excel on particular dimensions, none consistently performs well across all six dimensions in all three tasks, suggesting that comprehensive fitness AQA remains a challenge for current models.

\subsection{Controlled Perception-Oracle Evaluation}

We use paired perception and judgement questions to isolate how perception evidence affects judgement. The baseline asks the judgement question directly from the video. Question-only adds the paired perception question without its answer, while model-perception also supplies the model's answer. Comparing these settings separates any prompting effect of the perception question from the contribution of its answer. GT-perception instead provides the ground-truth answer, while GT-perception, no-video further removes the video. The judgement question remains unchanged across settings. Importantly, although judgement answers are derived from perception answers during dataset construction, the perception questions and options describe only neutral posture and motion evidence. Therefore, supplying the perception question and its answer only provides perception evidence, leaving the model to reason with domain knowledge.
As shown in Table~\ref{tab:oracle_perception}, question-only and model-perception bring little improvement over the baseline, whereas GT-perception substantially improves error detection. 
Notably, GPT-5.5 reaches 94.4\% F1 with GT-perception, much higher than Gemma-4-31B and Qwen3.5-27B. 
These results indicate that perception is a key bottleneck, especially for models with strong knowledge and reasoning capabilities.
Figure~\ref{fig:oracle_perception} further varies the correctness of the perception oracle from 0\% to 100\%, using incorrect answers at 0\%, ground-truth answers at 100\%, and mixtures in between. 
Judgement accuracy increases monotonically for all three models, showing a positive dependence of judgement performance on perception quality under controlled perception corruption.

\subsection{Temporal Grounding Failure Cases}

Figure~\ref{fig:grounding_vis} presents two representative temporal grounding failure cases. In the high-knees example, GPT-5.5 predicts almost the whole exercise segment, while the ground-truth interval only covers the early repetitions where the thighs are not raised high enough. This indicates that the model finds the action, but fails to pinpoint which repetitions contain the form error. In the butt-kicks example, the insufficient back-kick amplitude makes the erroneous repetitions deviate from the standard appearance of the exercise, so the model appears to treat them as not performing the target action and instead localizes later, cleaner repetitions. These cases suggest that the model relies heavily on coarse action semantics and struggles to identify quality-related deviations within the same exercise sequence.

\section{Conclusion}

We present FitAQA, a systematic benchmark for evaluating MLLMs in fitness action quality assessment. FitAQA contains 2,219 videos and 5,512 QA instances spanning 30 bodyweight exercises. It defines a unified taxonomy covering 38 form errors across six quality dimensions and includes three hierarchical evaluation tasks: perception, judgement, and temporal grounding. Experiments show that current MLLMs still struggle to assess exercise quality comprehensively, particularly in visual perception and temporal localization. Performance on judgement improves substantially when provided with perception evidence, identifying visual perception as a key bottleneck. These findings suggest that current MLLMs have yet to translate their strong semantic video understanding capabilities into reliable quality-sensitive action assessment. We believe that FitAQA provides a testbed for developing more reliable fitness AQA models.

\section*{Acknowledgement}

This research was backed by Huawei's AI Hundred Schools Program, and the experimental and computational work in this research leveraged AI compute resources alongside the Huawei Cloud AI Compute Service. 
This work was also supported by Capital's Funds for Health Improvement and Research (CFH) (2026-2-2191) and Beijing Natural Science Foundation (L242049).

\bibliography{aaai2027}

\clearpage
\appendix

\twocolumn[
\begin{@twocolumnfalse}
\begin{center}
{\LARGE\bfseries Appendix\par}
\vspace{1em}
\end{center}
\end{@twocolumnfalse}
]

\definecolor{tagcardio}{HTML}{E15759}
\definecolor{tagupper}{HTML}{6FA8DC}
\definecolor{taglower}{HTML}{7FBE7B}
\definecolor{tagcore}{HTML}{C783B9}
\definecolor{tagflex}{HTML}{E0B84F}
\newcommand{\tagbox}[1]{\raisebox{0.1ex}{\textcolor{#1}{\rule{0.75em}{0.75em}}}}
\newcommand{\tagCardio}{\tagbox{tagcardio}}
\newcommand{\tagUpper}{\tagbox{tagupper}}
\newcommand{\tagLower}{\tagbox{taglower}}
\newcommand{\tagCore}{\tagbox{tagcore}}
\newcommand{\tagFlex}{\tagbox{tagflex}}

\definecolor{promptbg}{RGB}{248,248,248}

\newtcblisting[auto counter]{promptbox}[2][]{
  enhanced,
  colback=promptbg,
  colframe=black!50,
  colbacktitle=black!50,
  coltitle=white,
  title={Prompt~\thetcbcounter: #2},
  fonttitle=\bfseries,
  boxsep=2pt,
  left=4pt,
  right=4pt,
  top=1.5pt,
  bottom=1.5pt,
  segmentation hidden,
  segmentation style={draw=none,fill=none},
  lower separated=false,
  listing only,
  listing options={
    basicstyle=\fontfamily{zi4}\fontsize{9}{10}\selectfont,
    breaklines=true,
    columns=fullflexible,
    keepspaces=true,
    showstringspaces=false,
    tabsize=2,
    numbers=none,
    xleftmargin=0pt,
    frame=none,
    aboveskip=0pt,
    belowskip=0pt
  },
  #1
}

\section{Dataset Construction Details}

\subsection{Action List}

Table~\ref{tab:action_list} lists the 30 target bodyweight exercises covered by FitAQA\@. These actions are selected to cover common home-based fitness scenarios and span five broad attributes: full-body cardio, upper-body strength, lower-body strength, core strength, and flexibility. Since some compound exercises involve multiple body regions or training goals, an action may be assigned to more than one attribute.

\begin{table}[htbp]
  \centering
  \small
  \begin{tabular}{lll}
    \toprule
    \textbf{No.} & \textbf{Attributes} & \textbf{Action Type} \\ \midrule
    1 & \tagCardio & Jumping Jack \\
    2 & \tagCardio & Quick Feet \\
    3 & \tagCardio & High Knees \\
    4 & \tagCardio & Butt Kicks \\
    5 & \tagCardio\,\tagLower & Tuck Jump \\
    6 & \tagCardio\,\tagLower & Jumping Lunge \\
    7 & \tagCardio\,\tagCore & Mountain Climbers \\
    8 & \tagCardio\,\tagUpper\,\tagLower\,\tagCore & Burpee \\
    9 & \tagUpper & Standing YTW \\
    10 & \tagUpper\,\tagCore & Knee Push-up \\
    11 & \tagUpper\,\tagCore & Push-up \\
    12 & \tagUpper\,\tagCore & Up-Down Plank \\
    13 & \tagLower & Side-Lying Leg Raise \\
    14 & \tagLower & Squat \\
    15 & \tagLower & Forward Lunge \\
    16 & \tagLower & Side Lunge \\
    17 & \tagLower & Vertical Jump \\
    18 & \tagLower\,\tagCore & Glute Bridge \\
    19 & \tagLower\,\tagCore & Single-Leg Glute Bridge \\
    20 & \tagCore & Dead Bug \\
    21 & \tagCore & Bird Dog \\
    22 & \tagCore & Reverse Crunch \\
    23 & \tagCore & Alternating Leg V-Up \\
    24 & \tagCore & Plank \\
    25 & \tagCore & Side Plank \\
    26 & \tagFlex & Cat-Cow Stretch \\
    27 & \tagFlex & Downward-Facing Dog \\
    28 & \tagFlex & Arm Circles \\
    29 & \tagFlex & Posterior Deltoid Stretch \\
    30 & \tagFlex & Overhead Triceps Stretch \\ \bottomrule
  \end{tabular}
  \caption{Action list. Colored squares denote attributes: \tagCardio~Full-Body Cardio; \tagUpper~Upper-Body Strength; \tagLower~Lower-Body Strength; \tagCore~Core Strength; \tagFlex~Flexibility.}
  \label{tab:action_list}
\end{table}

\subsection{Short-Clip Selection}

For fitness AQA datasets, we select videos from QEVD-FIT-300k and EgoExo-Fitness whose exercise names can be mapped to one of the 30 target actions. 
For generic action-recognition datasets, we first select fitness-related action classes to form a candidate video pool. From Kinetics-700, we select videos annotated as \textit{jumping jacks}, \textit{squat}, \textit{doing aerobics}, \textit{exercising arm}, \textit{exercising with an exercise ball}, \textit{high kick}, \textit{lunge}, and \textit{yoga}. From UCF101, we select videos from \textit{bodyweight squats}, \textit{jumping jack}, \textit{lunges}, and \textit{push ups}.
Since the class labels in Kinetics-700 and UCF101 are coarse-grained and may include videos with multiple people or irrelevant exercise variants, we further prompt Qwen3-VL-8B-Instruct to retain only candidate videos that contain one of the target exercises performed by a single visible subject, as shown in Prompt~\ref{prompt:short_clip_selection}.
We retain generic-dataset videos only when the model returns \texttt{keep=true} and the matched action belongs to the target action list. 
We then combine these retained generic-dataset clips with the selected clips from QEVD-FIT-300k and EgoExo-Fitness. This yields 2,168 short clips for human annotation.

\begin{figure}[t!]
\centering
\begin{promptbox}[label={prompt:short_clip_selection}]{Video Selection Prompt}
You are given a video and a target fitness action list. Please determine whether the video should be kept.

Keep the video only if all of the following conditions are satisfied:
1. The video contains one visible person.
2. The person is performing one action from the target action list.

Return your answer in the following JSON format:
{
  "keep": true or false,
  "matched_action": "one action from the target action list, or null"
}

<VIDEO>
Target action list: <TARGET_ACTION_LIST>
\end{promptbox}
\end{figure}

\subsection{Long-Video Selection}

For the long-video subset, we construct candidate videos from the longer-video subsets of QEVD. Since the original workout videos contain multiple exercises, we first use the start and end timestamps provided by the source dataset to segment them into single-exercise videos, resulting in 1,519 segments. We then map the exercise names of these segments to our 30 target actions and retain segments whose action type falls within the target action list, yielding 856 long-video segments for human annotation.

\subsection{Unified Form Error Taxonomy}

\begin{table*}[p!]
  \centering
  \footnotesize
  \begin{tabular}{@{}
    >{\raggedright\arraybackslash}p{0.13\textwidth}
    >{\centering\arraybackslash}p{0.03\textwidth}
    >{\raggedright\arraybackslash}p{0.28\textwidth}
    >{\raggedright\arraybackslash}p{0.48\textwidth}
    @{}}
    \toprule
    \textbf{Dimension} & \textbf{No.} & \textbf{Form Error} & \textbf{Example Manifestation} \\
    \midrule
    \multirow{25}{*}{I. Alignment\textsuperscript{*}}
    & 1 & Excessive Ankle Eversion & Medial arch collapse during vertical jump landing. \\
    & 2 & Insufficient Ankle Dorsiflexion & Heels lifted off the floor in squat or downward-facing dog. \\
    & 3 & Insufficient Knee Flexion & Knees stay too straight in quick feet. \\
    & 4 & Excessive Knee Flexion & Working leg bent in side-lying leg raise. \\
    & 5 & Excessive Hip Internal Rotation & Knees collapse inward during weight-bearing lower-body movements. \\
    & 6 & Excessive Hip External Rotation & Knees turned too far outward in squat or glute bridge. \\
    & 7 & Excessive Hip Abduction & Knee flares too far outward during leg drive in mountain climbers. \\
    & 8 & Excessive Hip Adduction & Knee crosses too far inward during leg drive in mountain climbers. \\
    & 9 & Insufficient Hip Flexion & Torso too upright from limited hip hinge in quick feet. \\
    & 10 & Excessive Hip Flexion & Torso leans too far forward in forward lunge, side lunge, or jumping lunge. \\
    & 11 & Insufficient Shoulder External Rotation & Elbows rotate inward during arm circles or standing YTW\@. \\
    & 12 & Insufficient Shoulder Internal Rotation & Elbow points forward instead of outward in overhead triceps stretch. \\
    & 13 & Insufficient Shoulder Flexion & Shoulder angle remains too closed in downward-facing dog support. \\
    & 14 & Excessive Shoulder Abduction & Elbows flared too wide from the torso in push-up. \\
    & 15 & Excessive Scapular Elevation & Shoulders shrugged toward the ears in plank or downward-facing dog. \\
    & 16 & Insufficient Scapular Protraction & Shoulder blades not spread enough in plank or push-up. \\
    & 17 & Excessive Elbow Flexion & Elbows too bent in posterior deltoid stretch or downward dog support. \\
    & 18 & Excessive Lumbar Flexion & Low back rounds in squat or push-up. \\
    & 19 & Excessive Lumbar Extension & Low back sags in push-up, bird dog extension, or reverse crunch lowering. \\
    & 20 & Excessive Thoracic Flexion & Rounded upper back in plank, push-up, or downward-facing dog. \\
    & 21 & Excessive Cervical Flexion & Head lifted with chin tucked in dead bug or reverse crunch. \\
    & 22 & Excessive Cervical Extension & Gaze lifted toward the ceiling instead of forward in squat. \\
    & 23 & Forward Head Posture & Head reaches forward toward the floor in push-up. \\
    & 24 & Incorrect Movement Trajectory & Arms drift forward out of the side plane during standing YTW transitions. \\
    & 25 & Incorrect Support Placement & Improper hand spacing in push-up or stance width in squat. \\
    \midrule
    \multirow{2}{*}{II. Symmetry}
    & 26 & Load Distribution Asymmetry & Uneven shoulder height during downward-facing dog support. \\
    & 27 & Range-of-Motion Asymmetry & Uneven leg height in high knees, or asymmetric limb lowering in dead bug. \\
    \midrule
    \multirow{2}{*}{III. Stability}
    & 28 & Trunk Instability & Trunk sways side to side with alternating legs in mountain climbers. \\
    & 29 & Support Instability & Unstable landing support in jumping lunge, or shaky leg in glute bridge. \\
    \midrule
    IV. Coordination
    & 30 & Movement Incoordination & Unsynchronized arm-leg motion in jumping jack, or same-side limbs used in dead bug. \\
    \midrule
    \multirow{3}{*}{V. Tempo}
    & 31 & Loss of Eccentric Tempo Control & Letting the leg and torso drop during alternating leg V-up descent. \\
    & 32 & Momentum-Driven Execution & Swinging the legs to complete reverse crunch. \\
    & 33 & Low Movement Cadence & Slow leg lift or long pause between steps in high knees. \\
    \midrule
    \multirow{5}{*}{VI. Completeness}
    & 34 & Insufficient Range of Motion & Limited arm raise, arm return, or leg spread in jumping jack. \\
    & 35 & Missing Movement Phase & Missing push-up or jump phase in burpee. \\
    & 36 & Insufficient Movement Duration & No hold after entering downward-facing dog. \\
    & 37 & Extraneous Movement & Extra heel lift during glute bridge ascent, or head lift during descent. \\
    & 38 & Exercise Mismatch & Performing substantially different from the requested exercise. \\
    \midrule
    \makecell[l]{Auxiliary Tags\textsuperscript{\dag}}
    & - & Incomplete Subject Visibility & Key body parts out of frame or occluded by scene objects. \\
    & - & Poor Lighting Condition & Body posture hard to assess due to dim or uneven lighting. \\
    \midrule
    \multicolumn{4}{@{}p{\textwidth}@{}}{\textsuperscript{*}Alignment errors capture postural deviations outside the primary movement range; insufficient amplitude of the main action is labeled as Insufficient Range of Motion rather than as a joint-specific alignment error (e.g., limited arm raise in jumping jack is not labeled as insufficient shoulder flexion).} \\
    \multicolumn{4}{@{}p{\textwidth}@{}}{\textsuperscript{\dag}These auxiliary video-quality tags are used to identify videos whose visual conditions make reliable form assessment difficult.} \\
    \bottomrule
  \end{tabular}
  \caption{Unified form error taxonomy and auxiliary video-quality tags with example exercise-specific manifestations.}
  \label{tab:form_error_taxonomy}
\end{table*}

Table~\ref{tab:form_error_taxonomy} provides the full unified form error taxonomy used for annotation. The taxonomy contains 38 form errors organized into six quality dimensions: alignment, symmetry, stability, coordination, tempo, and completeness. Each form error is accompanied by an example manifestation to illustrate how the same abstract error type can appear in concrete exercises. The table also includes two auxiliary video-quality tags, \textit{incomplete subject visibility} and \textit{poor lighting condition}, which are used for further video filtering.

The unified form error taxonomy was drafted by two experts based on their theoretical knowledge and practical experience in bodyweight fitness training, focusing on posture and motion errors that commonly recur across bodyweight exercises. 
During annotation, the taxonomy was further refined in an iterative manner: annotators reported ambiguous or uncovered cases, and the experts updated the taxonomy when a recurring error could not be cleanly assigned to an existing type.
For instance, \textit{missing movement phase} was added after annotators observed executions of compound actions such as burpees where a required phase was skipped entirely. 
After annotation, we removed error types that may be plausible for other bodyweight exercises but have no corresponding samples in our dataset, such as knee hyperextension, to ensure that the final taxonomy remains grounded in FitAQA.

\subsection{Annotation Protocol}

\subsubsection{Annotators}
The annotation involved two groups of annotators with distinct roles. The expert annotators were sports-science experts who collaborated with us on the benchmark construction. They drafted and updated the unified form error taxonomy, conducted pre-annotation training, reviewed the completed annotations, and revised the generated perception and judgement question pairs. The three student annotators were recruited from sports colleges based on their domain background and familiarity with common bodyweight exercises. They performed the initial video annotation, including selecting visible form-error labels, writing textual descriptions, marking auxiliary video-quality tags, and annotating temporal intervals for long videos.

\subsubsection{Training}
Before annotation, two experts introduced the correct execution standards and common form errors for the 30 target exercises. Annotators were trained with the unified form error taxonomy and example videos before annotating the benchmark data.

\subsubsection{Annotation Instructions}
Annotators were instructed to identify all visible form errors according to the unified taxonomy. They annotated only errors that could be confidently determined from the video. If camera viewpoint, frame rate, occlusion, or other visual limitations made an error uncertain, the error was not annotated. When the subject was partially outside the frame or the video was too dark to assess the full movement, annotators marked the corresponding video-quality tag and still annotated any other form errors that were visible.
Annotators followed additional disambiguation rules for labels that could otherwise be confused. They labeled \textit{exercise mismatch} only when the performed action fundamentally differed from the requested exercise in its core movement pattern, moving/static property, involved movement phase, movement direction, or action type, rather than for local posture errors or insufficient range of motion. When insufficient range of motion and local posture errors co-occurred, annotators labeled only \textit{insufficient range of motion} if the posture issue was a direct manifestation of the limited movement range. Otherwise, they labeled both errors.
For short clips, annotators selected all visible form-error labels and wrote a video-specific description for each selected label. The descriptions focused on concrete and observable visual evidence, as if explaining the video to someone who cannot see it.
For long videos, annotators performed the same label and description annotation, and additionally marked the temporal interval(s) where each described error occurred. When the same form-error type appeared with different manifestations over time, annotators wrote separate descriptions and marked the corresponding intervals for each unique description. Annotators also indicated whether the execution quality varied over time, which was later used to identify videos suitable for temporal-grounding evaluation.

\subsubsection{Annotation Tools}
\begin{figure*}[t!]
  \centering
  \includegraphics[width=\textwidth]{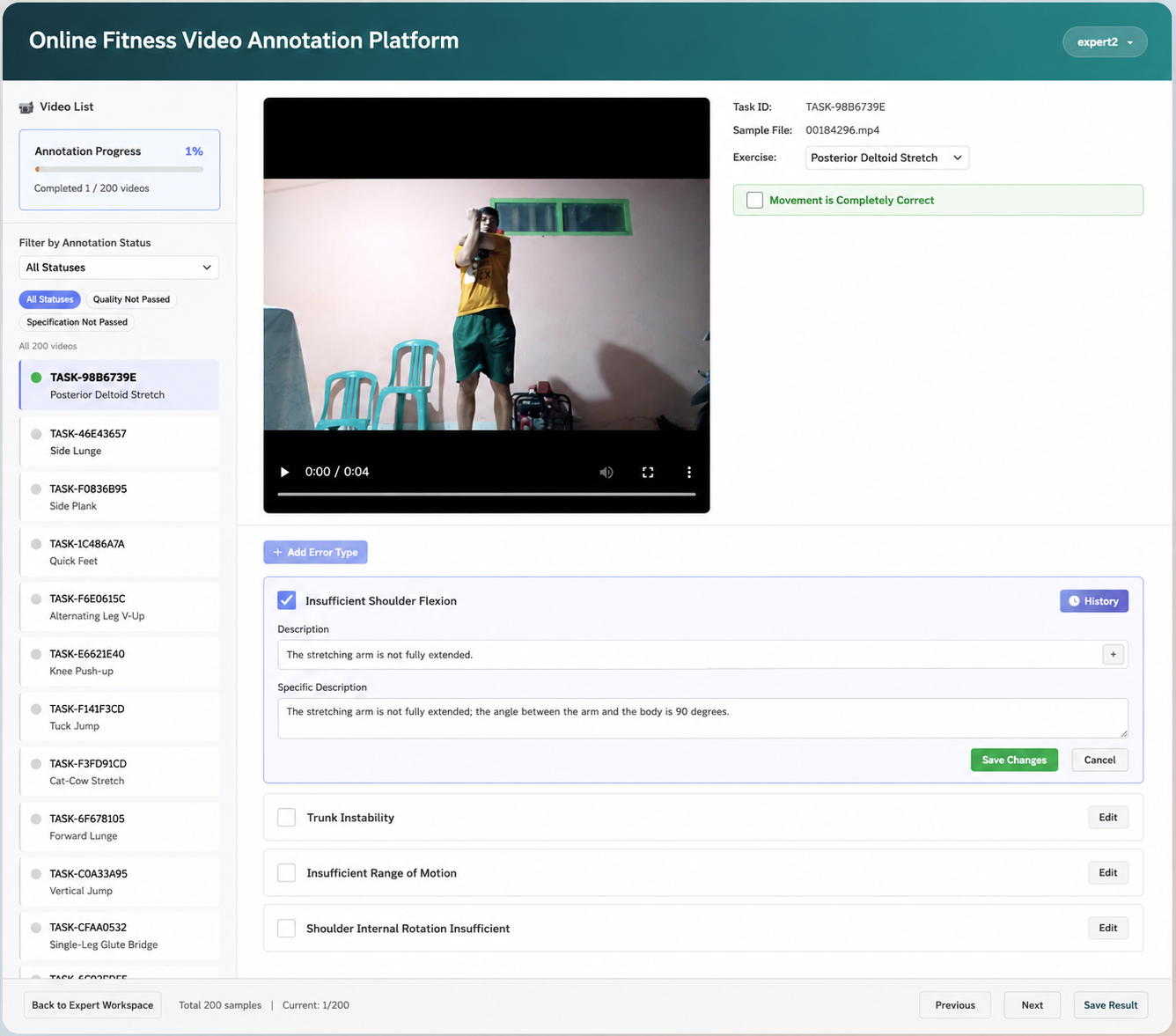}
  \caption{Screenshot of the custom-built web annotation system.}
  \label{fig:annotation_web}
\end{figure*}

For short clips, annotators used our custom-built web annotation system, as shown in Figure~\ref{fig:annotation_web}. For long videos, annotators used the video annotation interface of VGG Image Annotator~\cite{dutta2019via}, which supports temporal segment annotation in videos.

\subsubsection{Video Filtering}
For short clips, we filtered out videos whose only labels were \textit{exercise mismatch}, \textit{poor lighting condition}, or \textit{incomplete subject visibility}. Videos with these quality-related issues were retained only when they also contained recognizable form errors that could be assessed from the visible evidence. 
This step filtered out 192 videos, corresponding to 8.9\% of the annotated short-clip candidates.
For long videos, we retained only videos whose execution quality varied over time, removing clips that were consistently correct or consistently exhibited the same error pattern throughout the segment. 
This step filtered out 613 videos, corresponding to 71.6\% of the annotated long-video candidates.

\subsubsection{Expert Review}

\begin{table}[t!]
  \centering
  \small
  \begin{tabular}{lcc}
    \toprule
    \multirow{2}{*}{\textbf{Source}} &
    \multirow{2}{*}{\textbf{\#Videos}} &
    \textbf{Avg} \\
    & & \textbf{Duration} \\ \midrule
    \textit{Short-clip subset} & & \\
    QEVD-FIT-300k & 1,551 & 5.7 \\
    EgoExo-Fitness & 183 & 9.3 \\
    Kinetics-700 & 186 & 7.5 \\
    UCF101 & 56 & -- \\
    \textbf{Short-clip subtotal} & 1,976 & 6.2 \\ \midrule
    \textit{Long-video subset} & & \\
    QEVD-FIT-COACH & 83 & 33.2 \\
    QEVD-FIT-COACH-Benchmark & 44 & 33.2 \\
    QEVD-FIT-COACH-Competition & 116 & 26.5 \\
    \textbf{Long-video subtotal} & 243 & 30.0 \\ \midrule
    \textbf{Overall total} & 2,219 & -- \\ \bottomrule
  \end{tabular}
  \caption{Video composition of FitAQA. Average video duration is in seconds. UCF101 consists of image sequences.}
  \label{tab:video_composition}
\end{table}

Each video was first annotated by one student annotator and then reviewed by one expert. For retained videos, the expert checked whether the selected form-error labels, textual descriptions, and temporal intervals were consistent with the taxonomy and visible video evidence. Expert review led to annotation revisions for 162 short clips, corresponding to 8.2\% of the retained short clips, and 24 long videos, corresponding to 9.9\% of the retained long videos. For filtered-out samples, experts further checked 20 short clips and 62 long videos, accounting for approximately 10\% of the excluded samples in each subset. All sampled filtering decisions were confirmed to be correct.

\subsection{Video Composition}

After annotation and video filtering, FitAQA consists of 2,219 videos in total. The short-clip subset contains 1,976 clips from QEVD-FIT-300k, EgoExo-Fitness, Kinetics-700, and UCF101, and is used for perception and judgement evaluation. The long-video subset contains 243 videos from the longer-video subsets of QEVD, and is used for temporal grounding evaluation. Table~\ref{tab:video_composition} summarizes the final source-wise composition and average video duration. For UCF101, we do not report average duration because the retained samples are image sequences rather than standard video files.

\subsection{Question Generation}

For perception and judgement, we first group the annotated short clips by action type and form-error type, and collect the video-specific descriptions within each group. We then use GPT-5.4 to generate one perception question and one judgement question for each action-error group. This produces 348 initial perception-judgement question pairs. As shown in Prompt~\ref{prompt:question_generation}, the model is instructed to generate multiple-choice questions primarily grounded in the annotated descriptions and visual variations observed in the dataset.

\begin{figure*}[t!]
\centering
\begin{promptbox}[label={prompt:question_generation}]{Question Generation Prompt}
You are an expert in sports science. For a given error type of a specific exercise, design high-quality multiple-choice questions based on the textual descriptions of this error type across all video samples. The questions should distinguish differences in how different videos exhibit this error type.

{
  "action_name": <ACTION_NAME>,
  "total_num_videos": <ACTION_VIDEO_COUNT>,
  "form_error_type": <FORM_ERROR_TYPE>,
  "num_videos_without_this_error": <NUM_UNMARKED_VIDEOS>,
  "descriptions_of_this_error": <DESCRIPTIONS>
}

You need to generate one `perception` question and one `judgement` question.
* `perception`: The question should rely only on visual information in the video and should be answerable without knowing the standard for correct exercise execution. For example, "In the video, are the left and right hands raised to the same height?"
* `judgement`: The question should require both visual information and knowledge of proper exercise form to assess the error or movement quality. For example, "In the video, is the arm range of motion performed correctly?"

Requirements:
* `perception`: Each question should provide 2-4 options. The question stem and options should describe only visible phenomena in the video, without any subjective evaluation or technical terminology. Option A should correspond to the correct execution, while the other options should correspond to different types of incorrect execution. The options should be logically mutually exclusive with clear boundaries. Do not design questions for which multiple options may apply. Do not use vague degree words.
* `judgement`: The question should ask whether the movement is performed correctly with respect to the target aspect. Do not directly copy the name of the error type in the question; instead, convert it into a more natural expression. The options should be "Yes" and "No". Do not explain the judging criteria or visual characteristics associated with this error type.
* Both types of questions should be grounded in the video content. Do not ask purely theoretical questions.

Notes:
* Videos that are not annotated with this error type should be treated as correct in this aspect.
* You may add reasonable options based on the textual descriptions of existing samples, even if an option does not exactly correspond to any existing description.

Output schema:
```json
[
  {
    "question_type": "perception",
    "question": "question stem",
    "options": {
      "A": "option 1",
      "B": "option 2",
      "..."
    }
  },
  {
    "question_type": "judgement",
    "question": "judgement stem",
    "options": {
      "A": "Yes",
      "B": "No"
    }
  }
]
```

\end{promptbox}
\end{figure*}

\begin{figure}[t!]
\centering
\begin{promptbox}[label={prompt:answer_inference}]{Answer Inference Prompt}
You are an expert in sports science. Answer the given question based on the structured annotation of the video.

{
  "action_name": <ACTION_NAME>,
  "question": <QUESTION>,
  "form_error_type_of_the_question": <FORM_ERROR_TYPE>,
  "options": <OPTIONS>,
  "video_annotation": <FULL_VIDEO_ANNOTATION>
}

Notes:
* The structured annotation contains the error types that occur in the video and their corresponding textual descriptions. If an error type does not appear in the annotation, the video should be treated as correct in that aspect.
* If the full annotation uniquely supports one option, set `can_answer=true` and provide the corresponding option letter.
* If the video annotation does not contain sufficient information to answer the question, set `can_answer=false` and `answer=null`.

Output schema:
```json
{
  "can_answer": true,
  "answer": "B"
}
```

\end{promptbox}
\end{figure}

The generated question pairs are further revised by experts before answer inference. During this review, experts verify that the perception options remain neutral descriptions and that each option has an unambiguous semantic mapping to the matched judgement answer. Under this expert-verified mapping, the option associated with correct execution is indexed as A during construction and maps to \textit{Yes}, while the remaining options map to \textit{No}. Experts also merge question pairs when different form-error labels lead to the same judgement aspect. For instance, \textit{Excessive Hip Internal Rotation} and \textit{Excessive Hip External Rotation} can both be assessed through whether the knees track in the correct direction. Keeping separate judgement questions for such labels would make the answer depend on the generating error type rather than the actual queried aspect. We therefore merge them into one aspect-level question to ensure consistent answer assignment. By merging groups of two or three related question pairs, experts consolidate 110 initial question pairs into 50 question pairs.
For the remaining 238 non-merged question pairs, 42 pairs receive wording-level revisions, and another 9 pairs receive substantial rewrites. The wording-level revisions mainly adjust option phrasing to make the distinctions among options clearer while keeping the tested visual cue and answer mapping unchanged. Substantial rewrites are used when the original question and options are ambiguous or do not clearly capture the intended form-error aspect.

After revision, FitAQA contains 288 unique perception and judgement question pairs, which are then instantiated with videos to form the QA instances.

\subsection{Answer Inference}

We construct 20,491 candidate matched perception--judgement question pairs, corresponding to 40,982 QA instances. For each candidate perception instance, we use GPT-5.4-mini to infer the answer from the sample's full annotation. The prompt is shown in Prompt~\ref{prompt:answer_inference}. The model is allowed to abstain when the annotations do not provide sufficient evidence to select a unique option. We then derive the corresponding judgement answer from the perception answer: option A is mapped to \textit{Yes}, non-A options are mapped to \textit{No}, and abstained perception answers remain abstained.

After automatic inference, experts inspect the corresponding video to verify each inferred answer. They correct the option label when the video evidence supports a different option and mark an instance as unanswerable when the video does not support a unique answer. Any correction to a perception answer is propagated to the matched judgement answer through the same mapping. Table~\ref{tab:answer_verification} summarizes the changes made by experts during answer verification. Among 40,982 candidate QA instances, 1,196 change answerability status, yielding an answerability correction rate of 2.92\%. Specifically, 884 instances initially marked as unanswerable are assigned a valid answer, while 312 initially answerable instances are marked as unanswerable. Among the 37,270 instances that remain answerable before and after verification, experts correct the selected option for 1,044, corresponding to an answer correction rate of 2.80\%. Overall, 2,240 instances (5.47\%) receive either an answerability or option-label correction.
After verification, 38,154 of the 40,982 candidate QA instances remain answerable. Because each perception question has a matched judgement question, this total comprises 19,077 answerable perception QA instances and 19,077 matched judgement QA instances.

\begin{table}[t!]
  \centering
  \small
  \begin{tabular}{lcc}
    \toprule
    & \multicolumn{2}{c}{\textbf{Before verification}} \\
    \cmidrule(lr){2-3}
    \textbf{After verification} & Answerable & Unanswerable \\
    \midrule
    Answerable   & 37,270 (1,044) & 884 \\
    Unanswerable & 312            & 2,516 \\
    \bottomrule
  \end{tabular}
  \caption{Answer revisions during expert verification. Counts include matched judgement instances updated through the fixed perception-to-judgement mapping. The value in parentheses denotes the number of option-label corrections among instances that remained answerable.}
  \label{tab:answer_verification}
\end{table}

\subsection{MILP-Based Downsampling}

\begin{figure}[t!]
  \includegraphics[width=\columnwidth]{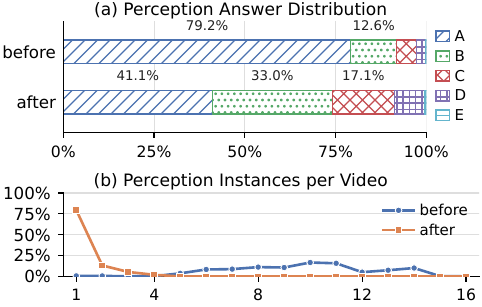}
  \caption{Effect of MILP-based downsampling on (a) perception answer options and (b) the number of perception instances per video.}
  \label{fig:balanced_downsampling}
\end{figure}

After verification, the perception instances remain strongly imbalanced across answer options. Because each question is applied to videos of the same exercise, many instances do not exhibit the queried form error, making the option associated with correct execution of the queried aspect dominant. Retaining all instances would allow benchmark scores to be driven by this frequent answer rather than reliable recognition of quality-relevant visual evidence. We therefore downsample the verified instances while preserving dataset coverage.
Specifically, we formulate the problem as a mixed-integer linear program (MILP). It requires every source video and every observed question-option bucket to remain represented, limits each video to four instances, and restricts overrepresented buckets relative to the other options for the same question. The objective further favors distributing retained instances across questions and answer options rather than repeatedly selecting from common buckets.

The procedure retains 2,562 perception QA instances and the same number of corresponding judgement instances. It preserves all 1,976 source videos and 288 perception questions. As shown in Fig.~\ref{fig:balanced_downsampling}(a), the option associated with correct execution accounts for 79.2\% of the verified perception instances before downsampling and 41.1\% afterward. Fig.~\ref{fig:balanced_downsampling}(b) further shows that no video contributes more than four retained instances. Overall, the procedure substantially reduces answer and video concentration while preserving the coverage of the verified candidate pool.

\section{Dataset Quality Validation}

We conduct a blinded cross-check on a stratified random sample covering the six quality dimensions. Within each dimension, we sample 10\% of the final perception and temporal grounding instances, with each sampled perception instance accompanied by its matched judgement instance. This yields 257 perception instances, 257 matched judgement instances, and 39 temporal grounding instances. During the cross-check, experts do not have access to the original answers or temporal intervals. Each expert independently re-annotates perception and temporal grounding instances previously reviewed by the other expert. Judgement answers are then derived from the re-annotated perception answers using the same mapping as in dataset construction.

For perception, the two experts agree on 253 of the 257 answers, yielding an exact-answer agreement of 98.4\%. Because judgement answers are mapped from perception answers, the mapped judgement answers inherit the same agreement. For temporal grounding, all intervals annotated by each expert are treated as a temporal set. We compute IoU from the total duration of the intersection and union of the two sets for each instance, obtaining an mIoU of 97.3\% across the 39 sampled instances. The cross-check is used only to measure agreement, and discrepancies do not alter the final annotations. The blinded cross-check provides evidence of high annotation consistency.

\begin{figure}[t!]
\centering
\begin{promptbox}[label={prompt:mcq_evaluation}]{Perception and Judgement Prompt}
You are an expert fitness video question answering assistant.
Answer the multiple-choice question based only on the given video.
Return exactly one option letter from the provided choices. Do not explain.

<VIDEO>

Action: <ACTION_NAME>
Question: <QUESTION>
Options:
<OPTION_LINES>

Answer with exactly one option letter.
\end{promptbox}
\end{figure}

\begin{figure}[t!]
\centering
\begin{promptbox}[label={prompt:temporal_grounding}]{Temporal Grounding Prompt}
You are given a fitness exercise video and one form-error description. Find the time interval(s) where the described error is clearly visible.

<VIDEO>

Exercise action: <ACTION_NAME>
Error description: <ERROR_DETAIL>

Return continuous interval(s), not individual frames. If the error occurs in multiple disjoint intervals, return multiple segments. Return only valid JSON, with no markdown or explanation. 

Use exactly this schema:
{"segments":[{"start":1.0,"end":7.5}]}
\end{promptbox}
\end{figure}

\begin{figure}[t!]
\centering
\begin{promptbox}[label={prompt:oracle_perception}]{Perception-Oracle Prompt Template}
You are an expert fitness video question answering assistant.
Answer the multiple-choice judgement question using the available evidence.
Return exactly one option letter from the provided choices. Do not explain.

<VIDEO>

Perception evidence:
Question: <PERCEPTION_QUESTION>
Options:
<PERCEPTION_OPTION_LINES>
Answer: <PERCEPTION_ANSWER>

Based on the above perception evidence, answer the following judgement question.

Action: <ACTION_NAME>
Question: <JUDGEMENT_QUESTION>
Options:
<JUDGEMENT_OPTION_LINES>

Answer with exactly one option letter.
\end{promptbox}
\end{figure}

\section{Evaluation Details}

\subsection{Evaluation Prompts}

For perception and judgement, we use the multiple-choice prompt shown in Prompt~\ref{prompt:mcq_evaluation}. During dataset construction, the answer options follow a fixed order, with option A in each perception question denoting the manifestation associated with correct execution. To remove positional cues, we shuffle the options of each QA instance once before evaluation and remap its ground-truth label accordingly. The resulting option order is fixed across all evaluated models and runs.

For temporal grounding, general-purpose MLLMs use the prompt shown in Prompt~\ref{prompt:temporal_grounding}.
For specialized temporal grounding models, we retain their official prompting formats and native temporal representations. All models receive the same exercise action and annotated form error description. Model-specific outputs, including textual time ranges, sampled frame indices, and discrete temporal tokens, are converted into intervals in seconds before evaluation. A single predicted interval is treated as a singleton temporal set.

\subsection{Controlled Perception-Oracle Prompts}

Prompt~\ref{prompt:oracle_perception} presents the prompt template used in the controlled perception-oracle evaluation. In the model-perception and GT-perception settings, \texttt{<PERCEPTION\_ANSWER>} is instantiated with the model-predicted and ground-truth perception answers, respectively. In the question-only setting, \texttt{<PERCEPTION\_ANSWER>} is omitted from the template. In the GT-perception with no video setting, \texttt{<VIDEO>} is omitted from the GT-perception prompt. The judgement question remains identical across all settings, and its ground-truth answer is never provided. All settings use the same fixed, pre-shuffled option order described above, with no reshuffling during inference.

\subsection{Output Parsing Rules}

\subsubsection{Perception and Judgement}

We strip leading and trailing whitespace from each response and normalize the extracted option label to uppercase. A response is valid only if exactly one label among the provided options can be identified. Responses containing no valid label or multiple distinct labels are treated as invalid. Invalid responses are counted as incorrect, and no additional language model is used to interpret or correct malformed outputs.

\subsubsection{Temporal Grounding}

For general-purpose MLLMs, we parse the required \texttt{segments} field. The native outputs of specialized models are first converted into the same list of intervals in seconds. We discard intervals with missing fields or non-finite \texttt{start} and \texttt{end} values. Valid timestamps are clipped to the video range \([0,T]\), where \(T\) is the video duration, and intervals with a non-positive duration after clipping are removed. The remaining intervals are sorted chronologically, and overlapping intervals are merged. If none remains, the response is treated as an empty prediction.

\subsection{Evaluation Metrics}

\subsubsection{Perception}

We report accuracy and question-macro accuracy (Q-MAcc). Let \(N_{\mathrm{p}}\) denote the number of perception instances, \(g_i\) the ground-truth option label, and \(\hat{g}_i\) the parsed prediction, with \(\hat{g}_i=\emptyset\) for an invalid response. Accuracy is defined as
\[
\operatorname{Acc}
=
\frac{1}{N_{\mathrm{p}}}
\sum_{i=1}^{N_{\mathrm{p}}}
\mathbf{1}[\hat{g}_i=g_i].
\]
Let \(\mathcal{Q}\) denote the set of unique perception questions and \(\mathcal{I}_q\) the set of instances associated with question \(q\). Q-MAcc is defined as
\[
\operatorname{Q\text{-}MAcc}
=
\frac{1}{|\mathcal{Q}|}
\sum_{q\in\mathcal{Q}}
\frac{1}{|\mathcal{I}_q|}
\sum_{i\in\mathcal{I}_q}
\mathbf{1}[\hat{g}_i=g_i].
\]
By assigning equal weight to each unique question, Q-MAcc prevents questions instantiated with more videos from dominating the aggregate score.

\subsubsection{Judgement}

We treat the answer indicating a form error as the positive class and report recall, precision, and F1 for this class. To ensure that invalid responses are always penalized, we count them as false negatives for error-positive instances and as false positives for error-negative instances. Using the resulting counts, the metrics are defined as
\[
\operatorname{Recall}
=
\frac{\mathrm{TP}}{\mathrm{TP}+\mathrm{FN}},
\qquad
\operatorname{Precision}
=
\frac{\mathrm{TP}}{\mathrm{TP}+\mathrm{FP}},
\]
\[
\operatorname{F1}
=
\frac{2\mathrm{TP}}
{2\mathrm{TP}+\mathrm{FP}+\mathrm{FN}}.
\]
A metric is set to zero when its denominator is zero.

\subsubsection{Temporal Grounding}

\newcommand{\meanstd}[2]{\ensuremath{#1_{\scriptscriptstyle #2}}}
\begin{table*}[t!]
  \centering
  \setlength{\tabcolsep}{2pt}
  \begin{tabular}{lc|cc|ccc|cccc}
    \toprule
    \multirow{2}{*}{\textbf{Model}} &
    \multirow{2}{*}{\textbf{Think}} &
    \multicolumn{2}{c|}{\textbf{Perception}} &
    \multicolumn{3}{c|}{\textbf{Judgement}} &
    \multicolumn{4}{c}{\textbf{Temporal Grounding}} \\
    & & \textbf{Accuracy} & \textbf{Q-MAcc} & \textbf{Recall} & \textbf{Precision} & \textbf{F1} & \textbf{R@0.3} & \textbf{R@0.5} & \textbf{R@0.7} & \textbf{mIoU} \\
    \midrule
    \multicolumn{11}{l}{\textit{Simple Baselines}} \\
    Uniform Random & -- & 33.6 & 35.7 & 50.0 & 58.9 & 54.1 & -- & -- & -- & -- \\
    Always Correct & -- & 41.1 & 43.2 & 0.0 & 0.0 & 0.0 & -- & -- & -- & -- \\
    Always Error & -- & -- & -- & 100.0 & 58.9 & 74.2 & -- & -- & -- & -- \\
    Whole Video & -- & -- & -- & -- & -- & -- & 25.5 & 4.9 & 0.0 & 23.2 \\
    \midrule
    \multicolumn{11}{l}{\textit{Temporal Grounding Models}} \\
    Grounded-VideoLLM-Phi3.5 & -- & -- & -- & -- & -- & -- & \meanstd{11.9}{1.5} & \meanstd{4.4}{0.8} & \meanstd{0.5}{0.3} & \meanstd{13.1}{1.6} \\
    VideoMind & -- & -- & -- & -- & -- & -- & 33.0 & 14.9 & 4.9 & 22.8 \\
    TimeLens & -- & -- & -- & -- & -- & -- & 28.6 & 16.2 & 7.5 & 22.0 \\
    \midrule
    \multicolumn{11}{l}{\textit{Open-source Models}} \\
    VideoLLaMA3-7B & \xmark & \meanstd{38.4}{0.6} & \meanstd{40.8}{0.5} & \meanstd{18.9}{1.1} & \meanstd{59.6}{0.8} & \meanstd{28.7}{0.9} & \meanstd{27.3}{1.8} & \meanstd{12.1}{1.5} & \meanstd{4.9}{0.9} & \meanstd{18.4}{1.6} \\
    InternVL3.5-8B & \xmark & \meanstd{40.1}{0.7} & \meanstd{43.7}{0.8} & \meanstd{19.5}{1.0} & \meanstd{64.2}{0.9} & \meanstd{29.9}{0.9} & \meanstd{19.8}{2.1} & \meanstd{3.6}{0.7} & \meanstd{0.5}{0.3} & \meanstd{17.3}{1.8} \\
    InternVL3.5-30B-A3B & \xmark & \meanstd{41.7}{0.4} & \meanstd{44.6}{0.5} & \meanstd{10.3}{1.2} & \meanstd{65.1}{0.7} & \meanstd{17.7}{1.0} & \meanstd{15.2}{1.9} & \meanstd{5.2}{0.8} & \meanstd{1.3}{0.5} & \meanstd{12.1}{1.4} \\
    Gemma-4-31B & \xmark & \meanstd{43.4}{0.5} & \meanstd{46.3}{0.4} & \meanstd{17.6}{1.4} & \meanstd{65.8}{0.6} & \meanstd{27.8}{1.1} & \meanstd{35.6}{1.7} & \meanstd{16.5}{1.5} & \meanstd{5.4}{1.1} & \meanstd{24.9}{1.6} \\
    Qwen3-VL-8B & \xmark & \meanstd{41.3}{0.8} & \meanstd{44.2}{0.7} & \meanstd{16.9}{1.3} & \meanstd{64.9}{0.9} & \meanstd{26.8}{1.1} & \meanstd{12.4}{1.5} & \meanstd{3.4}{0.6} & \meanstd{0.3}{0.2} & \meanstd{10.1}{1.1} \\
    Qwen3-VL-30B-A3B & \xmark & \meanstd{41.4}{0.4} & \meanstd{44.3}{0.5} & \meanstd{12.1}{1.1} & \meanstd{64.8}{0.7} & \meanstd{20.3}{1.0} & \meanstd{32.7}{2.2} & \meanstd{15.7}{1.6} & \meanstd{7.5}{1.2} & \meanstd{24.3}{1.9} \\
    Qwen3.5-9B & \xmark & \meanstd{41.3}{0.7} & \meanstd{44.6}{0.6} & \meanstd{35.8}{1.4} & \meanstd{63.0}{1.2} & \meanstd{45.7}{1.3} & \meanstd{36.1}{1.8} & \meanstd{19.1}{1.4} & \meanstd{7.0}{1.0} & \meanstd{25.7}{1.7} \\
    Qwen3.5-9B & \cmark & \meanstd{41.1}{0.6} & \meanstd{43.9}{0.7} & \meanstd{28.9}{1.2} & \meanstd{64.5}{1.0} & \meanstd{39.9}{1.1} & \meanstd{42.3}{2.3} & \meanstd{22.7}{1.8} & \meanstd{10.3}{1.4} & \meanstd{30.4}{2.1} \\
    Qwen3.5-27B & \xmark & \meanstd{42.5}{0.5} & \meanstd{45.2}{0.4} & \meanstd{37.4}{1.3} & \meanstd{63.4}{0.9} & \meanstd{47.1}{1.1} & \meanstd{51.3}{1.9} & \meanstd{35.1}{1.7} & \meanstd{19.3}{1.5} & \meanstd{37.6}{1.8} \\
    Qwen3.5-27B & \cmark & \meanstd{42.6}{0.4} & \meanstd{45.7}{0.5} & \meanstd{31.3}{1.1} & \meanstd{66.7}{0.8} & \meanstd{42.6}{1.0} & \meanstd{52.3}{2.0} & \meanstd{30.2}{1.6} & \meanstd{17.0}{1.3} & \meanstd{36.6}{1.9} \\
    Qwen3.5-35B-A3B & \xmark & \meanstd{42.4}{0.5} & \meanstd{45.6}{0.4} & \meanstd{22.5}{1.4} & \meanstd{64.5}{0.9} & \meanstd{33.4}{1.2} & \meanstd{43.3}{1.7} & \meanstd{26.8}{1.5} & \meanstd{13.7}{1.2} & \meanstd{30.5}{1.6} \\
    Qwen3.5-35B-A3B & \cmark & \meanstd{42.5}{0.4} & \meanstd{45.9}{0.5} & \meanstd{27.7}{1.2} & \meanstd{65.7}{0.8} & \meanstd{39.0}{1.1} & \meanstd{45.9}{1.8} & \meanstd{25.8}{1.4} & \meanstd{12.9}{1.1} & \meanstd{31.7}{1.7} \\
    \midrule
    \multicolumn{11}{l}{\textit{Closed-source Models}} \\
    Gemini-3.1-pro-preview & \cmark & \meanstd{50.7}{0.3} & \meanstd{54.3}{0.4} & \meanstd{80.6}{0.9} & \meanstd{65.2}{0.6} & \meanstd{72.1}{0.8} & \meanstd{62.4}{1.6} & \meanstd{43.8}{1.5} & \meanstd{25.8}{1.3} & \meanstd{44.0}{1.4} \\
    \quad \textit{no video} &  & \meanstd{39.4}{1.0} & \meanstd{41.2}{1.0} & \meanstd{18.2}{0.7} & \meanstd{58.9}{0.9} & \meanstd{27.7}{0.8} & \meanstd{27.5}{1.8} & \meanstd{7.1}{0.4} & \meanstd{1.1}{0.7} & \meanstd{20.1}{0.4} \\
    GPT-5.4 & \cmark & \meanstd{47.6}{0.4} & \meanstd{51.3}{0.5} & \meanstd{38.8}{1.1} & \meanstd{72.3}{0.8} & \meanstd{50.5}{1.0} & \meanstd{51.3}{1.9} & \meanstd{28.5}{1.6} & \meanstd{18.6}{1.4} & \meanstd{36.9}{1.8} \\
    GPT-5.5 & \cmark & \meanstd{51.2}{0.3} & \meanstd{54.1}{0.4} & \meanstd{48.3}{1.0} & \meanstd{71.9}{0.7} & \meanstd{57.8}{0.9} & \meanstd{67.3}{1.7} & \meanstd{42.4}{1.5} & \meanstd{27.0}{1.3} & \meanstd{47.5}{1.6} \\
    \quad \textit{no video} &  & \meanstd{40.7}{0.4} & \meanstd{43.9}{0.7} & \meanstd{80.6}{1.2} & \meanstd{59.2}{0.3} & \meanstd{68.3}{0.6} & \meanstd{7.8}{0.5} & \meanstd{2.4}{0.8} & \meanstd{0.2}{0.3} & \meanstd{7.3}{0.2} \\
    \bottomrule
  \end{tabular}
  \caption{Main results across three runs, including the no-video controls used to analyze potential textual bias. Standard deviations are shown as subscripts where applicable. VideoMind and TimeLens use deterministic decoding in their official inference settings and are reported without standard deviations. All values are percentages.}
  \label{tab:main_results_across_runs}
\end{table*}

For each temporal grounding instance \(i\), let \(P_i\) and \(G_i\) denote the unions of all predicted and ground-truth intervals, respectively. The temporal intersection over union (tIoU) is defined as
\[
\operatorname{tIoU}_i
=
\frac{\mu(P_i\cap G_i)}
{\mu(P_i\cup G_i)},
\]
where \(\mu(\cdot)\) denotes the total duration of a temporal region. This set-based formulation naturally supports multiple disjoint intervals while ensuring that overlapping temporal regions are counted only once.

We report the mean IoU (mIoU), computed as the average tIoU over the \(N_{\mathrm g}\) temporal grounding instances:
\[
\operatorname{mIoU}
=
\frac{1}{N_{\mathrm g}}
\sum_{i=1}^{N_{\mathrm g}}
\operatorname{tIoU}_i.
\]

We additionally report Recall@\(\tau\), defined as the proportion of instances whose tIoU is at least \(\tau\):
\[
\operatorname{Recall@}\tau
=
\frac{1}{N_{\mathrm g}}
\sum_{i=1}^{N_{\mathrm g}}
\mathbf{1}\!\left[\operatorname{tIoU}_i\geq\tau\right].
\]
We use thresholds \(\tau\in\{0.3,0.5,0.7\}\).

\section{Additional Results}

\subsection{Main Results Across Runs}

Table~\ref{tab:main_results_across_runs} reports the mean and standard deviation over three runs where applicable. For completeness, we additionally report an always-error baseline, which predicts every execution as erroneous. Although it achieves an F1 of 74.2\%, this strategy is clearly unsuitable for practical fitness assessment. Overall, the results are stable across runs, with small standard deviations on most metrics. Although the relative ordering of similarly performing models varies in some cases, this variation does not affect our main conclusion that current MLLMs remain limited on FitAQA.

\subsection{Textual Bias Analysis}

\begin{figure*}
  \centering
  \includegraphics[width=0.95\textwidth]{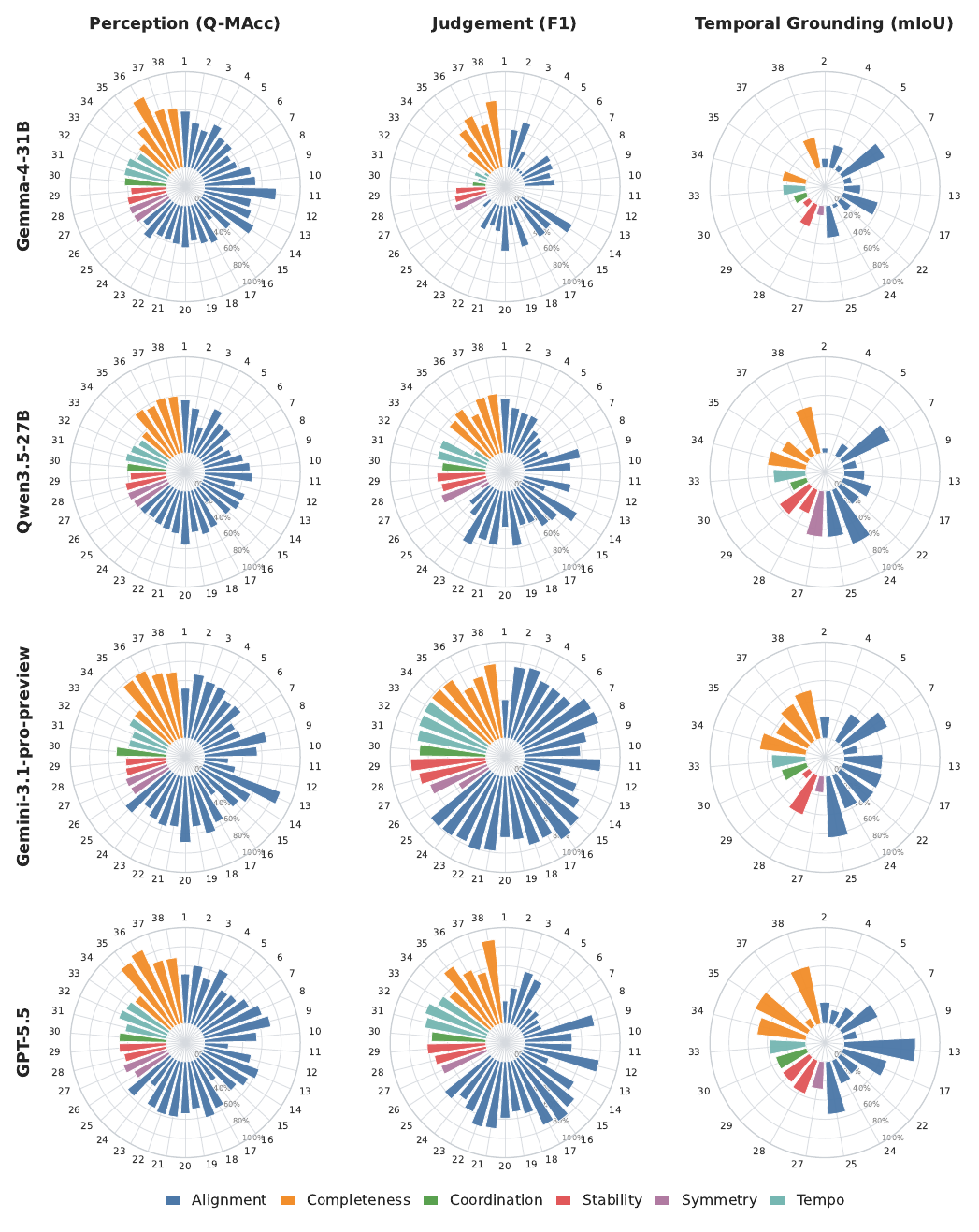}
  \caption{Results by form error for Gemma-4-31B, Qwen3.5-27B, Gemini-3.1-pro-preview, and GPT-5.5. The numbers around each polar plot indicate the corresponding taxonomy IDs in Table~\ref{tab:form_error_taxonomy}. Questions associated with multiple form errors contribute to the result for each corresponding error.}
  \label{fig:form_error_polar}
\end{figure*}

We additionally evaluate Gemini-3.1-pro-preview and GPT-5.5 without video input to probe potential textual bias. The results are reported in Table~\ref{tab:main_results_across_runs}. This setting retains all task-specific textual inputs and, for temporal grounding, provides the total video duration to define the valid timestamp range.
Removing the video substantially degrades perception and temporal grounding performance for both models. Perception Q-MAcc drops close to the always-correct baseline, while Recall@0.7 falls to nearly zero despite access to the total video duration. These results show that perception and temporal grounding relies on recognizing visual evidence rather than on textual priors.
Judgement exhibits a different pattern. Gemini-3.1-pro-preview deteriorates sharply without video, whereas GPT-5.5 achieves a higher F1 by predicting form errors much more frequently. Their no-video error recalls are 18.2\% and 80.6\%, respectively. Without visual evidence, Gemini tends to predict correct execution, whereas GPT-5.5 tends to predict form errors in the absence of visual evidence. Therefore, the relatively high no-video F1 of GPT-5.5 reflects a prediction bias toward the error class rather than reliable quality assessment.

\subsection{Results by Form Error}

Figure~\ref{fig:form_error_polar} shows that model performance varies substantially across individual form errors. In perception, GPT-5.5 performs particularly well on ID 7 (Excessive Hip Abduction) and ID 8 (Excessive Hip Adduction), while Gemini-3.1-pro-preview achieves the highest Q-MAcc on ID 13 (Insufficient Shoulder Flexion). ID 16 (Insufficient Scapular Protraction) remains difficult for all four models, with Q-MAcc below 29\%. For judgement, Gemini-3.1-pro-preview shows a clear advantage on ID 5 (Excessive Hip Internal Rotation), ID 6 (Excessive Hip External Rotation), ID 7 (Excessive Hip Abduction), and ID 8 (Excessive Hip Adduction), whereas ID 26 (Load Distribution Asymmetry) is consistently challenging.
The temporal grounding results reveal complementary strengths. Qwen3.5-27B performs best on ID 24 (Incorrect Movement Trajectory), Gemini-3.1-pro-preview on ID 25 (Incorrect Support Placement) and ID 37 (Extraneous Movement), and GPT-5.5 on ID 13 (Insufficient Shoulder Flexion) and ID 35 (Missing Movement Phase). Moreover, the strongest model for a given form error is not always consistent across perception, judgement, and temporal grounding.

\subsection{Paired Perception and Judgement Results}

Table~\ref{tab:joint_perception_judgement} examines the joint correctness of paired perception and judgement questions. GPT-5.5 achieves the highest proportion of pairs for which both answers are correct (40.1\%). Gemini-3.1-pro-preview, however, has the lowest rate of joint failure, with only 19.4\% of pairs answered incorrectly on both questions.
The asymmetric cases further reveal that perception and judgement are related but not equivalent capabilities. For Gemini-3.1-pro-preview, 29.8\% of all paired instances fall into the perception-incorrect and judgement-correct category, a larger proportion than for the other models. More specifically, its judgement accuracy remains 60.6\% among perception-incorrect pairs, compared with 65.7\% among perception-correct pairs. This relatively small difference shows that an incorrect answer about the specific visual manifestation does not necessarily lead to an incorrect binary judgement. In contrast, Gemma-4-31B exhibits a much stronger coupling between the two tasks: its judgement accuracy is 78.7\% when perception is correct but only 21.0\% when perception is incorrect.
Conversely, all four models exhibit cases in which perception is correct but judgement is incorrect, ranging from 9.2\% for Gemma-4-31B to 17.4\% for Gemini-3.1-pro-preview. Thus, correctly recognizing the relevant visual state does not always translate into correctly assessing whether the movement constitutes a form error. Overall, these results indicate that perception provides useful evidence for judgement, while successful judgement additionally requires interpreting the observed state against domain knowledge.

\subsection{Perception and Judgement Cases}

\begin{figure*}
  \centering
  \includegraphics[width=\textwidth]{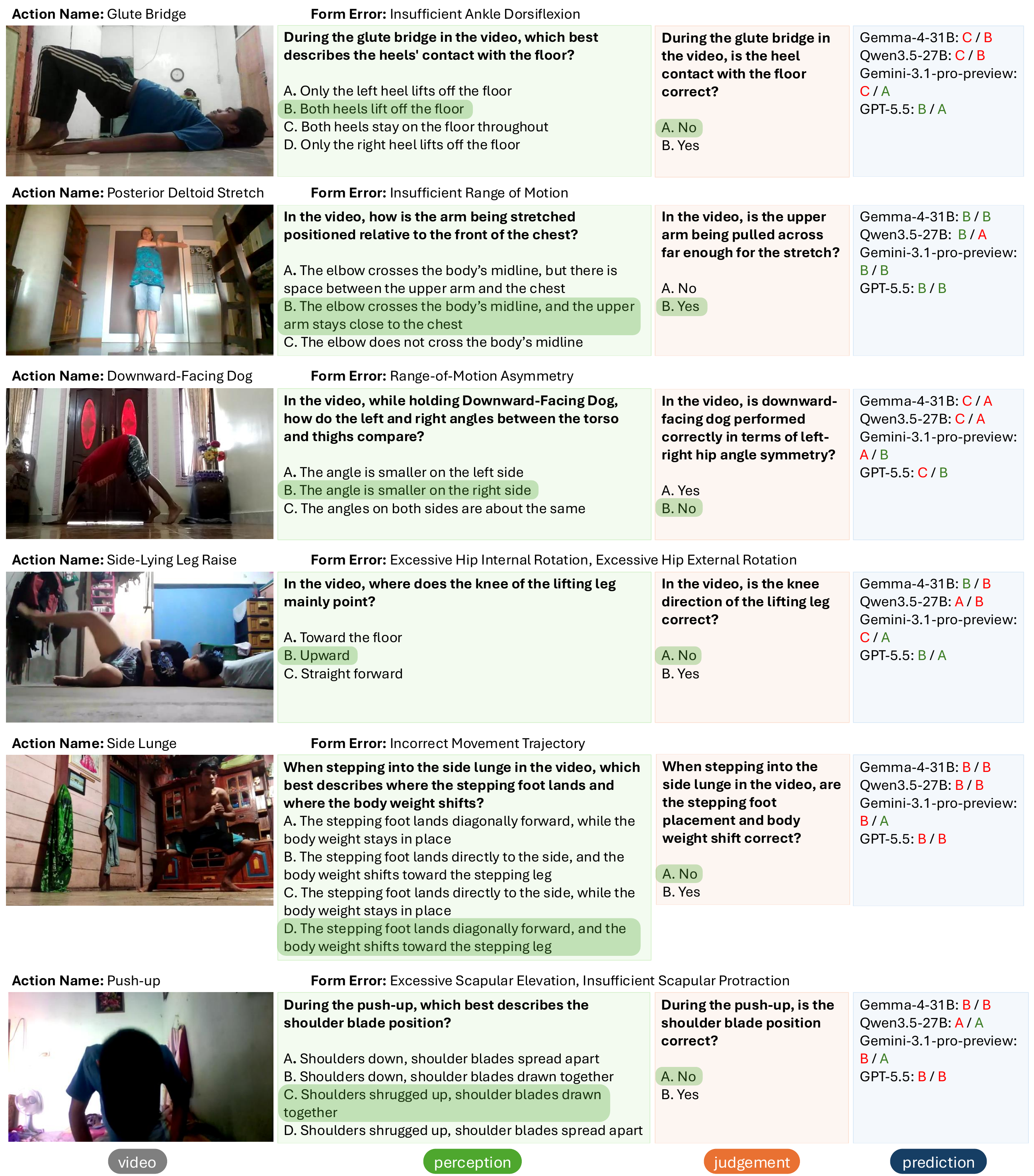}
  \caption{Representative paired perception and judgement QA instances. Ground-truth options are highlighted with green shading. Model outputs are shown as perception/judgement predictions, with green and red text denoting correct and incorrect predictions, respectively.}
  \label{fig:mcq_examples}
\end{figure*}

Figure~\ref{fig:mcq_examples} qualitatively illustrates the joint outcome patterns in Table~\ref{tab:joint_perception_judgement} through cases covering all four combinations of perception and judgement correctness. The examples in which perception is correct but judgement is incorrect show that a model may identify the observed visual manifestation yet assess it incorrectly against the exercise standard. Conversely, several correct judgement predictions accompany an incorrect perception answer. Because judgement is binary, such cases can conceal failures to identify the precise visual manifestation. These examples reinforce the value of paired evaluation for tracing errors to visual evidence recognition or subsequent quality judgement.

\subsection{Temporal Duration Distributions}

\begin{figure*}[t!]
  \centering
  \includegraphics[width=\textwidth]{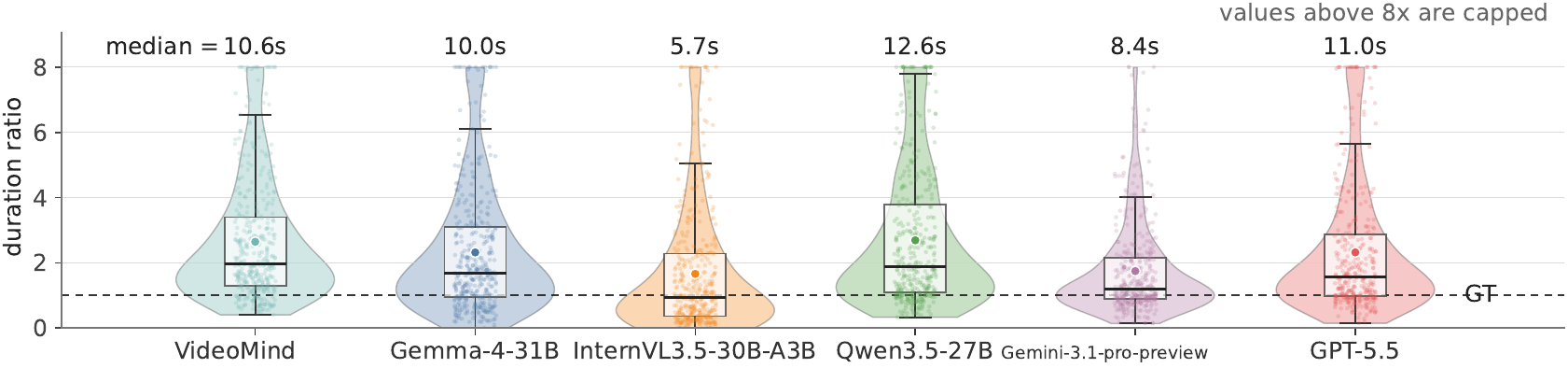}
  \caption{Distributions of the ratios between predicted and ground-truth temporal grounding durations. The values above each model report the median predicted duration. The median ground-truth duration is 5.7 seconds.}
  \label{fig:temporal_duration_distributions}
\end{figure*}

Figure~\ref{fig:temporal_duration_distributions} shows a common tendency toward temporal over-localization among the six representative models. Each distribution is constructed from the ratio of total predicted duration to total ground-truth duration for each instance, where a ratio above one indicates an overlong prediction. Five of the six models exhibit a median ratio above one, indicating that predictions longer than the annotated error regions are common. For GPT-5.5, these ratios average 2.43, while the predicted and ground-truth durations average 12.36 and 6.93 seconds, respectively. Such over-localization introduces excessive temporal coverage and reduces tIoU. This tendency is consistent with weak performance at stricter thresholds.

\subsection{Effect of Video Sampling Rate}

\begin{table}[t!]
  \centering
  \small
  \resizebox{\columnwidth}{!}{
  \begin{tabular}{lcccc}
    \toprule
    \textbf{Model} &
    \textbf{P\cmark J\cmark} &
    \textbf{P\cmark J\xmark} &
    \textbf{P\xmark J\cmark} &
    \textbf{P\xmark J\xmark} \\
    \midrule
    Gemma-4-31B   & 34.2 &  9.2 & 11.9 & 44.7 \\
    Qwen3.5-27B   & 28.4 & 14.1 & 21.9 & 35.6 \\
    Gemini-3.1-pro-preview  & 33.4 & 17.4 & 29.8 & 19.4 \\
    GPT-5.5       & 40.1 & 11.0 & 18.3 & 30.6 \\
    \bottomrule
  \end{tabular}}
  \caption{Joint correctness of paired perception (P) and judgement (J) predictions. \cmark{} and \xmark{} denote correct and incorrect predictions, respectively. All values are percentages.}
  \label{tab:joint_perception_judgement}
\end{table}

\begin{table}[t!]
  \centering
  \small
  \begin{tabular}{c|c|c|cc}
    \toprule
    \multirow{2}{*}{\textbf{FPS}} &
    \multirow{2}{*}{\makecell{\textbf{Perception}\\\textbf{Q-MAcc}}} &
    \multirow{2}{*}{\makecell{\textbf{Judgement}\\\textbf{F1}}} &
    \multicolumn{2}{c}{\textbf{Temporal Grounding}} \\
    & & & \textbf{R@0.7} & \textbf{mIoU} \\ \midrule
    1 & 44.3 & 50.4 & 9.8 & 30.3 \\
    2 & 45.1 & 47.0 & 18.0 & 36.0 \\
    4 & 45.6 & 41.8 & 19.9 & 37.6 \\
    8 & 46.4 & 43.1 & 18.3 & 35.9 \\ \bottomrule
  \end{tabular}
  \caption{Effect of video sampling rate on Qwen3.5-27B in non-thinking mode.}
  \label{tab:effect_of_fps}
\end{table}

Table~\ref{tab:effect_of_fps} evaluates Qwen3.5-27B in non-thinking mode with frame sampling rates ranging from 1 to 8 FPS. Perception Q-MAcc increases only modestly, from 44.3\% to 46.4\%, while judgement F1 decreases overall, from 50.4\% to 43.1\%. Temporal grounding improves up to 4 FPS, reaching 19.9\% Recall@0.7 and 37.6\% mIoU, before declining at 8 FPS. For Qwen3.5-27B, increasing the sampling rate alone does not consistently improve performance, suggesting that sparse sampling is not the sole limitation in this setting.

\begin{table}[t!]
  \centering
  \small
  \setlength{\tabcolsep}{4pt}
  \begin{tabular}{@{}
    >{\raggedright\arraybackslash}p{0.22\linewidth}
    >{\raggedright\arraybackslash}p{0.73\linewidth}
    @{}}
    \toprule
    \textbf{Source Dataset} & \textbf{Access and Redistribution} \\
    \midrule
    QEVD
    & Available from Qualcomm under the Data License Agreement--Research Use. We do not redistribute the videos; users must obtain QEVD from Qualcomm and comply with the agreement. \\
    \midrule
    EgoExo-Fitness
    & Available by request under the EgoExo-Fitness License Agreement. We do not redistribute the videos; users must request access from the dataset administrators and comply with the agreement. \\
    \midrule
    Kinetics-700
    & Available through the official Kinetics/CVDF distribution. We do not redistribute the videos; users should obtain them from the official source and follow the applicable terms. \\
    \midrule
    UCF101
    & Available through the official UCF CRCV dataset page. We do not redistribute the videos; users should obtain them from the official source and follow the applicable terms. \\
    \bottomrule
  \end{tabular}
  \caption{Source datasets used to construct FitAQA and their access and redistribution conditions.}
  \label{tab:artifact_terms}
\end{table}

\section{Limitations}
FitAQA currently focuses on bodyweight fitness exercises. While this setting covers many common movements, it does not include equipment-based exercises. Therefore, errors related to equipment usage, such as grip, load selection, equipment placement, or external-object trajectory, are not evaluated in the current version. The long-video subset is also smaller and less diverse than the short-clip subset. This is mainly because existing fitness datasets contain relatively few long videos with temporally varying execution quality, which are necessary for meaningful temporal grounding evaluation. Nevertheless, it provides a setting for temporal grounding evaluation, and future versions can expand it with more exercises, recording conditions, and error transitions.

\section{Ethical Considerations}
FitAQA is intended for research evaluation of MLLMs, rather than direct deployment as an automated fitness coaching, clinical rehabilitation, or medical decision-making system without further validation. Incorrect feedback may lead to inappropriate training guidance or injury risk; therefore, models evaluated on FitAQA should not replace professional advice without human oversight. FitAQA is built from existing third-party fitness and video datasets, and its release will respect the access, license, and redistribution conditions of the original sources. Since the videos may contain visible human subjects, we do not collect or annotate any new personally identifying information, and we release only QA instances, video metadata, and evaluation scripts rather than the original videos.

\section{Artifact Release and Terms of Use}

Table~\ref{tab:artifact_terms} summarizes the source datasets used to construct FitAQA, together with their access and redistribution conditions. 
We do not redistribute any original videos. Users must obtain the source datasets from their official providers and comply with the corresponding licenses or terms of use.

\end{document}